\documentclass{article}

\usepackage{microtype}
\usepackage{graphicx}
\usepackage{subfigure}
\usepackage{booktabs} % for professional tables
\usepackage{paralist}
\usepackage{amsmath}
\usepackage{amssymb}
\usepackage{csquotes}
\usepackage{xcolor}
\usepackage[section]{placeins}
\usepackage{colortbl}

\usepackage{hyperref}

\newcommand{\expect}{\mathop{\mathbb{E}}}

\newcommand{\ie}{\textit{i.e.},}
\newcommand{\eg}{\textit{e.g.},}

\newcommand{\vs}{\textit{vs.}}

\newcommand{\effect}{reward inflation}
\newcommand{\Effect}{Reward Inflation}

\DeclareMathOperator*{\argmax}{arg\,max}

\usepackage[accepted]{icml2021}

\begin{document}

\twocolumn[
\icmltitle{Do You Need the Entropy Reward (in Practice)?}




\begin{icmlauthorlist}
\icmlauthor{Haonan Yu}{hobot}
\icmlauthor{Haichao Zhang}{hobot}
\icmlauthor{Wei Xu}{hobot}
\end{icmlauthorlist}

\icmlaffiliation{hobot}{Horizon Robotics, Cupertino, CA, USA}

\icmlcorrespondingauthor{Haonan Yu}{haonan.yu@horizon.ai}

%\icmlkeywords{MaxEnt RL, entropy}

\vskip 0.3in
]



\printAffiliationsAndNotice{}  % leave blank if no need to mention equal contribution

\begin{abstract} %Soft actor-critic (SAC)~\citep{Haarnoja2018}, as a representative of 
Maximum entropy (MaxEnt) RL maximizes a combination of the original task reward and an entropy reward. 
It is believed that the regularization imposed by entropy, on both policy improvement and policy evaluation, together contributes to good exploration, training convergence, and robustness of learned policies.
This paper takes a closer look at entropy as an intrinsic reward, by conducting various 
ablation studies on soft actor-critic (SAC), a popular representative of MaxEnt RL. 
Our findings reveal that 
in general, entropy rewards should be applied with caution to policy evaluation. On one hand, the entropy reward, like any other intrinsic reward, could obscure the 
main task reward if it is not properly managed. We identify some failure cases of the entropy reward especially in episodic 
Markov decision processes (MDPs), where it could cause the policy to be overly optimistic or pessimistic. 
On the other hand, our large-scale empirical study shows that using entropy regularization alone in policy improvement, leads to comparable or even better performance and robustness than using it in both policy improvement and policy evaluation.
Based on these observations, we recommend either normalizing the entropy reward to a zero mean (SACZero), or simply 
removing it from policy evaluation (SACLite) for better practical results.
\end{abstract}

\section{Introduction}
\label{sec:introduction}

Since its appearance, soft actor-critic (SAC)~\citep{Haarnoja2018} has achieved a great success as one of the 
state-of-the-art continuous control reinforcement learning (RL) algorithms. 
Compared to other contemporary RL algorithms 
such as PPO~\citep{schulman2017proximal}, DDPG~\citep{Lillicrap2016}, and TD3~\citep{fujimoto2018addressing}, 
SAC enjoys the best of both worlds: off-policy training for sample efficiency and a stochastic actor for 
training stability. 
One major design of SAC is to augment the task reward with an entropy reward, resulting in a maximum entropy (MaxEnt) RL formulation.
Some primarily claimed benefits of MaxEnt RL are:
\begin{compactenum}[$\circ$]
    \item encouraging the agent to visit unexplored states where the policy entropy is expected to be high \citep{Haarnoja2018},
    \item smoothing the optimization landscape to make the optimization easier~\citep{ahmed2019understanding} and improving the         training convergence and stability~\citep{vieillard2020leverage}, 
    \item encouraging the agent to accomplish a task while acting randomly, 
        for robustness when deployed with environment perturbations \citep{eysenbach2021maximum}.
\end{compactenum}
%
%Benefits (i) and (ii) can typically be reflected by training performance while benefit (iii) requires an adverserial evaluation.

Compared to some prior works that merely use entropy as a regularization cost in policy improvement~\citep{mnih2016asynchronous, schulman2017proximal, abdolmaleki2018maximum}, 
using entropy to regularize \emph{both} policy improvement and evaluation seems more ``aggressive'' in 
transforming the original RL objective. 
%
%Although it has been shown that MaxEnt RL is robust 
%to both dynamics and reward perturbations~\citep{eysenbach2021maximum}, 
%
%RL practitioners have mostly benchmarked SAC on
%traditional control tasks (\eg\ MuJoCo locomotion~\citep{Brockman2016}) that emphasize little or no 
%evaluation robustness. 
%
Thus a natural question is, how largely will entropy as an intrinsic reward obscure the maximization 
of the utility? % which is usually what SAC is used for in the literature? 
Moreover, will robustness to environment 
perturbations also emerge (empirically) when entropy is only used for regularizing policy improvement?

One non-negligible issue of the entropy reward arising particularly in episodic Markov decision processes (MDPs) is that, 
it will change the policy's tendency of terminating an episode.
The reason is that entropy bonuses are added to normal time steps but not to
those post termination. 
This could result in overly optimistic or pessimistic policies, depending on whether the entropy is positive or negative.
In this paper, we name this side effect as \emph{\effect{}}.
Previously, \effect{} has been largely ignored by practitioners especially when applying SAC
to episodic MDPs. 
%
%Even for infinite-horizon MDPs, we conjecture that a similar side effect still 
%exists to some extent, due to the initial bias present in functionally approximated target Q values.

Another scenario which entropy rewards could complicate is multi-objective RL. 
There, an interplay between several task rewards already exists in the training process, and 
it is often difficult to determine a good trade-off among them. 
Having an entropy reward on top of multiple task rewards further makes
the training dynamics more unpredictable.

Of course, as pointed out by \citet{Haarnoja2018}, when the entropy weight $\alpha$ is sufficiently small (compared to the 
task reward scale), its MaxEnt RL objective reduces back to the original utility objective.
When $\alpha$ is tunable, this usually won't happen until the policy becomes much deterministic, due to a high initial value of $\alpha$\footnote{If $\alpha$ is already very small when training starts, or it's quickly tuned down 
to a small value, the agent's exploration ability will be impacted. The official SAC implementation %(\url{https://github.com/rail-berkeley/softlearning/blob/master/softlearning/algorithms/sac.py})
hardcodes the initial $\alpha$ to be $1.0$ and suggests a learning rate of $3\times 10^{-4}$.}.
It is best not to ignore the distortion of the utility objective by simply assuming that $\alpha$ will 
eventually become very small, as the training might have been affected in an irreversible way by then.

With these doubts in mind, we conduct ablation studies centering around the entropy that 
seems critical to SAC.
Through extensive experiments on various continuous control tasks, our findings reveal that in general, 
entropy rewards should be applied with caution to policy evaluation.
On one hand, we show that the entropy reward could sometimes obscure the task reward if its weight 
is not properly managed.
Particularly, we show manifestations of \effect{} that hinders utility maximization especially in episodic MDPs.
On the other hand, we show that entropy regularization alone in policy improvement leads to comparable or even better performance and robustness.
The gain of having the entropy bonus in policy evaluation on top of this is little or sometimes negative.
As a result, we recommend simply removing entropy from policy evaluation while only 
keeping it in policy improvement for better practical results (dubbed as SACLite). 
When the entropy reward is indeed needed, normalizing it to a zero mean will alleviate the \effect{} effect (dubbed as SACZero).
\ifdefined\isaccepted
The code for reproducing the experiment results is available at: \url{https://github.com/hnyu/entropy_reward}.
\else
The code for reproducing the experiment results will be made public.
\fi

\textbf{Disclaimer} This paper is not arguing that the entropy-augmented return formulation in MaxEnt RL is unnecessary.
%
%As already shown in \citet{eysenbach2021maximum} and \citet{vieillard2020leverage}, MaxEnt RL provably optimizes lower bounds on several robust objectives, 
%resulting in certain robustness to perturbations in the dynamics and reward.
%
Instead, it serves to bring to RL practitioners' attention that there is a hidden cost of using an entropy reward in policy evaluation, 
from a perspective of intrinsic rewards.
To minimize this cost, we propose to apply zero-mean normalization to the entropy reward, or better remove it from the return.
While SAC is studied as a representative MaxEnt RL algorithm, some conclusions could also 
apply to other MaxEnt RL algorithms, for example, \citet{haarnoja2017softq,nachum2017bridging,schulman2018equivalence}.

%More importantly, this cost is usually higher than the one introduced by merely 
%using entropy to regularize policy improvement.

\section{Related Work}
\textbf{Entropy regularized RL}\ \ Incorporating entropy regularization into RL has proved to be an effective technique for improving performance, especially in the deep RL literature~\citep{mnih2016asynchronous,schulman2017proximal,abdolmaleki2018maximum,haarnoja2017softq,Haarnoja2018}.
In contrast to using entropy for only regularizing policy improvement, the MaxEnt RL formulation~\citep{Ziebart2010} also uses it to regularize policy evaluation, by maximizing a combination of the task reward and the entropy reward.

To our best knowledge, there aren't many empirical studies that compare entropy regularizing policy improvement to entropy regularizing both policy improvement and evaluation.
%
%The former is usually treated as an implementation trick while the latter is thought to be more principled because its actual MDP is well defined under the framework of MaxEnt RL.
%
On a small subset of six Atari games, \citet{schulman2018equivalence} compared the two entropy use cases with an A2C backbone, where they call the two ``naive'' and ``proper'' versions of entropy-regularized RL, respectively. 
However, their experiment results were not very conclusive: the proper version was shown to be ``the same or possibly better than'' the naive version. 
On only CartPole and the Asterix Atari game, \citet{vieillard2020leverage} investigated if adding entropy rewards in
policy evaluation improves empirical results.
There was no significant benefit observed for Asterix.
For CartPole, the advantage of regularizing policy evaluation was only seen on certain hyperparameters.
In contrast to these two prior works, this paper analyzes the entropy regularization from a perspective of intrinsic rewards, and presents a large-scale study on complex control tasks.

\textbf{Intrinsic rewards}\ \ are internally calculated by RL agents independently of the extrinsic environment rewards.
They are usually for improving exploration by driving an agent to visit novel states~\citep{pathak2017curiositydriven,burda2018exploration} or learning diverse skills~\citep{Eysenbach2019,Gehring2021} in an unsupervised way.
Usually, intrinsic rewards are not exactly aligned with extrinsic rewards, and thus how to make a good trade-off between them is important~\citep{Badia2020}.
The entropy reward can be regarded as a special case of intrinsic rewards, where it drives an agent to visit states on which its policy produces actions randomly and thus is usually thought to be under-trained.

\section{Background}
%
%Our problem definition and notation generally follow the ones defined in SAC~\citep{Haarnoja2018}. 
%
The RL problem can be defined as policy search in an MDP $(\mathcal{S},\mathcal{A},p,r)$.
The state space $\mathcal{S}$ and action space $\mathcal{A}$ are both assumed to be continuous.
The environment transition probability $p(s_{t+1}|s_t,a_t)$ stands for the probability density of arriving 
at state $s_{t+1}\in \mathcal{S}$ after taking an action $a_t\in\mathcal{A}$ at the current state $s_t\in\mathcal{S}$.
It is usually unknown to the agent.
For each transition, the environment outputs a reward to the agent as $r(s_t,a_t,s_{t+1})$.
Sometimes, it is easier to just write the (expected) reward of the transition as $r(s_t,a_t)\triangleq\expect_{s_{t+1}\sim p(\cdot|s_t,a_t)}r(s_t,a_t,s_{t+1})$.
We use $\pi(a_t|s_t)$ to denote the agent's policy.

The typical RL objective is to find a policy $\pi^*$ that maximizes the expected discounted return 
\[\pi^*=\argmax_{\pi} \expect_{\pi,p}\sum_{t=0}^{\infty}\gamma^t r(s_t,a_t),\]
where $\gamma\in[0,1]$ is a discount factor.
For episodic settings, we assume $r(s_t,a_t)=0$ for time steps post episode termination.
The MaxEnt objective \citep{Ziebart2010,Haarnoja2018} augments this objective with an entropy bonus, as an intrinsic reward, 
so that the optimal policy maximizes not only  the task reward $r$ but also its own entropy:
\begin{equation*}
    \begin{array}{rl}
    \pi^*&=\displaystyle\argmax_{\pi} \expect_{\pi,p}\sum_{t=0}^{\infty}\gamma^t \big(r(s_t,a_t) + \alpha\gamma \mathcal{H}(\pi(\cdot|s_{t+1}))\big)\\
    &\triangleq\displaystyle\argmax_{\pi}\expect_{\pi,p}\sum_{t=0}^{\infty}\gamma^t r_{\text{MaxEnt}}(s_t,a_t),\\
    \end{array}
\end{equation*}
where $\alpha$ is the entropy weight. 
The soft Q value is defined as %$Q^{\pi}_{\text{MaxEnt}}(s_t,a_t)=r_{\text{MaxEnt}}(s_t,a_t)+\expect_{\pi,p}\sum_{t'=t+1}^{\infty}\gamma^{t'-t}r_{\text{MaxEnt}}(s_{t'},a_{t'})$,
$Q^{\pi}_{\text{MaxEnt}}(s_t,a_t)=\expect_{\pi,p}\sum_{t'=t}^{\infty}\gamma^{t'-t}r_{\text{MaxEnt}}(s_{t'},a_{t'})$,
representing the expected discounted return staring from $(s_t,a_t)$ and following $\pi$ thereafter.

\textbf{Soft actor-critic}~\citep{Haarnoja2018}.\ \ For practical implementations in an off-policy actor-critic setting, SAC parameterize $\pi$ and 
$Q^{\pi}_{\text{MaxEnt}}$ with $\phi$ and $\theta$, respectively. 
It then optimizes a surrogate objective 
\begin{equation}
\label{eq:policy_improvement}
\max_{\phi}\expect_{s\sim\mathcal{D},a\sim\pi_{\phi}}\big[Q_{\theta}(s,a)+\alpha \mathcal{H}(\pi_{\phi}(\cdot|s))\big],
\end{equation}
where $\mathcal{D}$ represents a replay buffer that stores past explored transitions. 
Meanwhile, $Q_{\theta}$ is learned by temporal-difference (TD) with the following Bellman backup operator
\begin{equation}
\label{eq:policy_evaluation}
\begin{array}{l}
\mathcal{T}^{\pi_{\phi}}Q_{\theta}(s_t,a_t)\\
\triangleq \displaystyle r_{\text{MaxEnt}}(s_t,a_t)+\gamma \expect_{\substack{s_{t+1},a_{t+1} \sim p, \pi_{\phi}}} Q_{\theta}(s_{t+1},a_{t+1}).\\
\end{array}
\end{equation}

As selecting a fixed entropy weight $\alpha$ is usually difficult without knowing the task reward scale beforehand, 
SAC automatically tunes it given an entropy target $\bar{\mathcal{H}}$ by solving 
\begin{equation}
\label{eq:alpha}
\min_{\alpha>0}\expect_{s\sim\mathcal{D}}\alpha(\mathcal{H}(\pi_{\phi}(\cdot|s))-\bar{\mathcal{H}}),
\end{equation}
together with Eq.~\ref{eq:policy_improvement} constitutes a minimax optimization problem.
Depending on the value of $\bar{\mathcal{H}}$, $\alpha$ typically decreases from a large initial value to 
a small one so that $\mathcal{H}(\pi_{\phi}(\cdot|s))$ is anchored to $\bar{\mathcal{H}}$ in expectation. 
This process makes the policy switch from exploration to exploitation gradually. 
%
%For more details about SAC, we refer the reader to \citet{Haarnoja2018}.

\textbf{Training with an episode time limit.}\ \ In RL training, it is a convention to truncate an unfinished episode by a time limit in case the agent gets stuck in a dead loop or uninteresting experiences. 
When an episode ends because of timeout and the time information is absent from the observation, the discount of the final step should be $\gamma$ instead of $0$. 
Thus even though an episode is truncated, value learning continues beyond the episode boundary.
In other words, whether a task is episodic or infinite-horizon is totally determined by the MDP instead of by how an algorithm truncates each episode for a training purpose.
In the remainder of this paper, both episodic and infinite-horizon MDPs could be defined with a time limit which is largely unrelated to our entropy reward discussions and analyses.

\section{Entropy Cost and Entropy Reward}
\label{sec:entropy_cost_and_reward}
Before our investigation starts, we'd like to first define two concepts to be used frequently in the remainder of this paper.
%
%They have already appeared earlier in our discussion:
%
\begin{compactenum}[i)]
    \item using entropy as a regularization term for policy improvement (Eq.~\ref{eq:policy_improvement}), and
    \item adding entropy values as intrinsic rewards (Eq.~\ref{eq:policy_evaluation}) to policy evaluation, 
        just like using other novelty-based rewards~\citep{pathak2017curiositydriven,burda2018exploration}.
\end{compactenum}
When referring to ``entropy regularized'' RL, most works don't differentiate in depth between the impacts of entropy on training 
performance from (i) and from (ii).
For the former, entropy serves as a regularization term that affects only one step of policy 
optimization~\citep{schulman2018equivalence}, while for the latter entropy behaves as an intrinsic reward 
that does reward shaping to the original MDP. 
In the following, we will use the term \emph{entropy cost} to refer to (i), while using \emph{entropy reward/bonus}
to refer to (ii). 
The occurrences of ``entropy reward'' prior to this section all refer to (ii).

It is tempting to treat the entropy cost as a one-step entropy bonus, for example in \citet{schulman2018equivalence}.
Although they are very similar, we believe that it is better not to confuse the two.
Generally speaking, a one-step entropy bonus depends on the next state $s_{t+1}$ and it is different for each action $a_t$ to be taken at 
the current state $s_t$, namely, it can reward actions differently.
This can be achieved by setting the discount $\gamma$ of any $s_{t+k}$ $(k>1)$ to zero.
In contrast, the entropy cost regularizes the policy at $s_t$ independent of specific $a_t$, namely, without differentiating the actions. 

\section{\Effect{} by the Entropy Reward}
For now let us focus on interpreting entropy as an intrinsic reward (ii).
Given the augmented reward $r_{\text{MaxEnt}}$, one would instantly ask the question: how 
will the entropy reward impact the policy learning? 
As with any intrinsic reward, it inevitably transforms the original MDP.
We know that it certainly requires the policy to make a trade-off between utility maximization and stochasticity. 
However, in this section we highlight a side effect orthogonal to stochasticity which we call the \emph{\effect{}}
of the entropy reward.
\subsection{A Toy Example in an Episodic Setting}
\begin{figure}[t!]
\begin{center}
\resizebox{\columnwidth}{!}{
\begin{tabular}{@{}c@{}|@{}c@{}}
    \includegraphics[width=0.5\columnwidth]{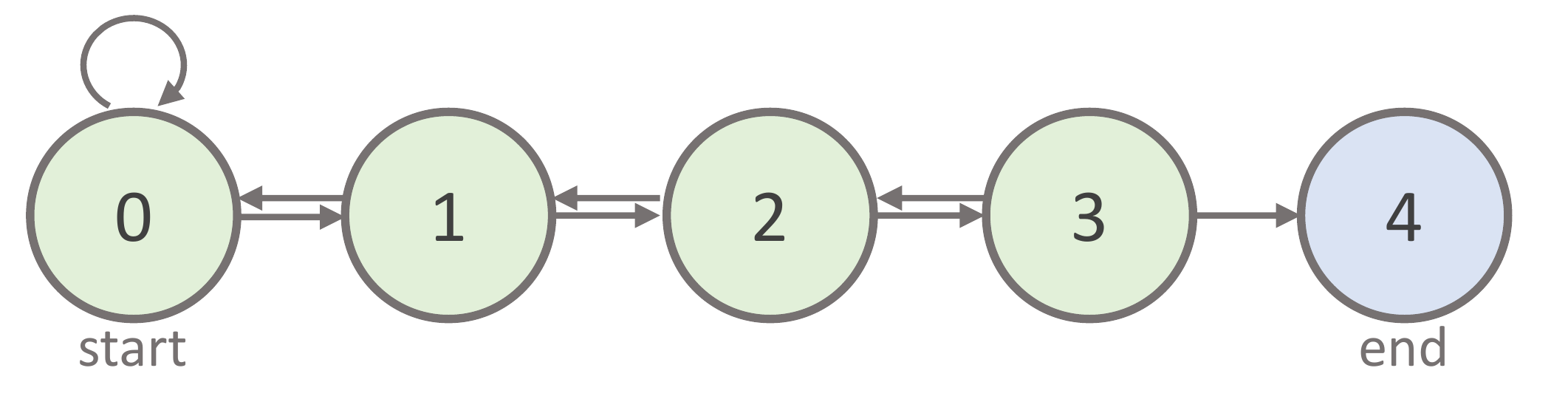}&
    \includegraphics[width=0.5\columnwidth]{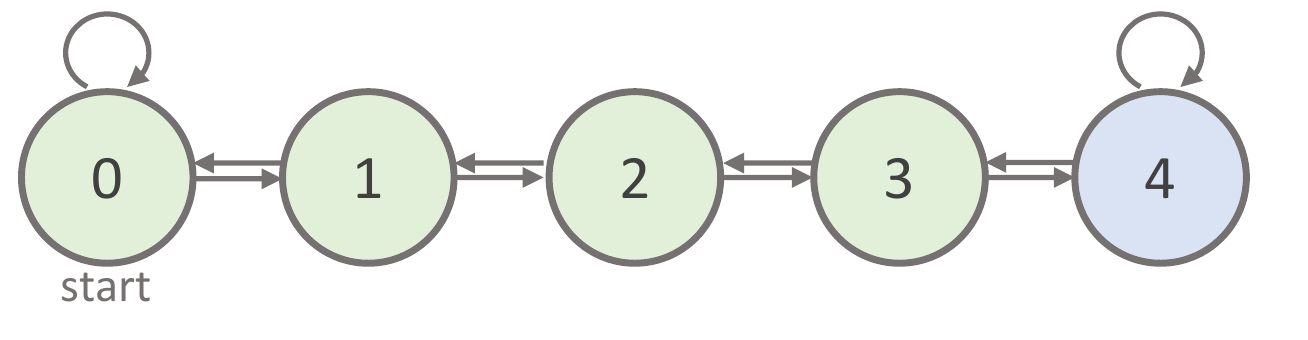}\\
\end{tabular}
\vspace{-4ex}}
    \caption{The \textsc{SimpleChain} tasks with a chain length of 5, with 
    $r(\text{green node})=-0.05$ and $r(\text{blue node})=0$.
    Taking a left (right) action at node 0 (4) makes the agent stay.
    Left: the episodic setting; Right: the infinite-horizon setting.
    }
\label{fig:simple_chain}
\end{center}
\vspace{-5ex}
\end{figure}

We will first demonstrate that how an entropy reward hinders policy learning on a toy task \textsc{SimpleChain}  
in an episodic setting (Figure~\ref{fig:simple_chain}, left). 
The environment consists of a simple chain of length 5. 
For each episode, the agent always starts from the leftmost node 0.
At each time step, it can choose to move left or right, and receives a reward of $-0.05$ for node 0-3.
The only way to end an episode is to reach the rightmost node 4 which gives a reward of $0$.
The time limit of an episode is set to 50 steps.
The agent's action space is $[-1,1]$, and its action in $[-1,0.8)$ is mapped to ``left'' and 
$[0.8,1]$ to ``right''.
Given this space, a uniform continuous action distribution has an entropy of $\log(2)\approx 0.7$.

If we train SAC using a fixed entropy weight $\alpha$ of $0.2$, in the beginning the agent will get an entropy reward 
of $\alpha\mathcal{H}(\pi)=0.2\times \log(2)\approx 0.14$ at every step.
So the overall reward it gets at a non-terminal state is roughly $0.14-0.05=0.09$. 
Thus due to entropy rewards, the agent's policy becomes overly optimistic and never wants to reach the terminal state.
When $\alpha$ is tunable and starts with $0.2$, as Eq.~\ref{eq:alpha} takes effect, if the entropy target is high, 
$\alpha\mathcal{H}(\pi)$ can still be greater than $0.05$ in its stable state 
(\eg\ $\alpha$ and $\mathcal{H}(\pi)$ settle at slightly smaller values than $0.2$ and $\log(2)$, respectively).    
Even for a low entropy target, the policy's tendency of termination will not flip until $\alpha\mathcal{H}(\pi)$ 
decreases below $0.05$.
This process might take some time depending on $\alpha$'s learning rate, and thus hurt the sample efficiency.

It is important to note that there exist many more training configurations where the entropy reward makes training fail.
This toy example is not special or restrictive; the absolute values of $-0.05$, $\log(2)$, and $0.2$ are not important. 
It is the relative magnitudes of these numbers that contribute to the behavior of the entropy reward.
For example, one can come up with another similar example where the agent has an $N$-dimensional action space and the immediate 
reward is $-0.05N$, or $\alpha=1.0$ with an immediate reward of $-0.1$.

\subsection{\Effect{} and Its Remedies}
\label{sec:leveling_effect}
The key issue SAC faces in the above is that its entropy reward inflates (in expectation) the reward function before termination while not after it, because the Q value of any step post termination is always set to zero when doing TD backup. 
Since an entropy reward could also be negative for a continuous distribution, it can also deflate (in expectation)
the reward function\footnote{An opposite behavior may happen if the entropy is negative initially (\eg\, when the action range is small and the uniform probability density is greater than $1$). In this case, the agent tends to terminate the task earlier than the optimal behavior which might be staying alive.}.
%
%\footnote{The deflation effect will be usually rare, because when entropy is negative, its weight $\alpha$ is expected to be so small that their product is negligible. However, this product is also decided by the action dimensionality and boundaries.}.
%
For terminology simplicity, we will use ``inflation'' to refer to both cases.
Reward inflation can cause a policy to be overly optimistic or pessimistic\footnote{Generally speaking, any intrinsic reward could have this \effect{} in an episodic setting. Here, we only focus on the entropy reward as a special case.}.

One aggressive way of reducing the \effect{} is to simply discard the entropy reward but still keeping the entropy cost in 
policy optimization (Eq.~\ref{eq:policy_improvement}).
This usage of entropy has previously appeared in other actor-critic algorithms, for example A3C~\citep{mnih2016asynchronous}, 
A2C~\citep{schulman2018equivalence}, and MPO~\citep{abdolmaleki2018maximum}.
In the following, we denote this variant of SAC by \emph{SACLite}.
Another remedy is to make sure that the entropy reward have a running mean of zero. 
Thus when doing TD backup in Eq.~\ref{eq:policy_evaluation}, we propose to subtract the entropy rewards by their moving average (\ie\ zero-mean normalization), resulting in a variant of SAC which we call \emph{SACZero}.

\begin{figure}[t!]
\begin{center}
\resizebox{0.8\columnwidth}{!}{
    \begin{tabular}{@{}c@{}c@{}}
        \includegraphics[width=0.45\columnwidth]{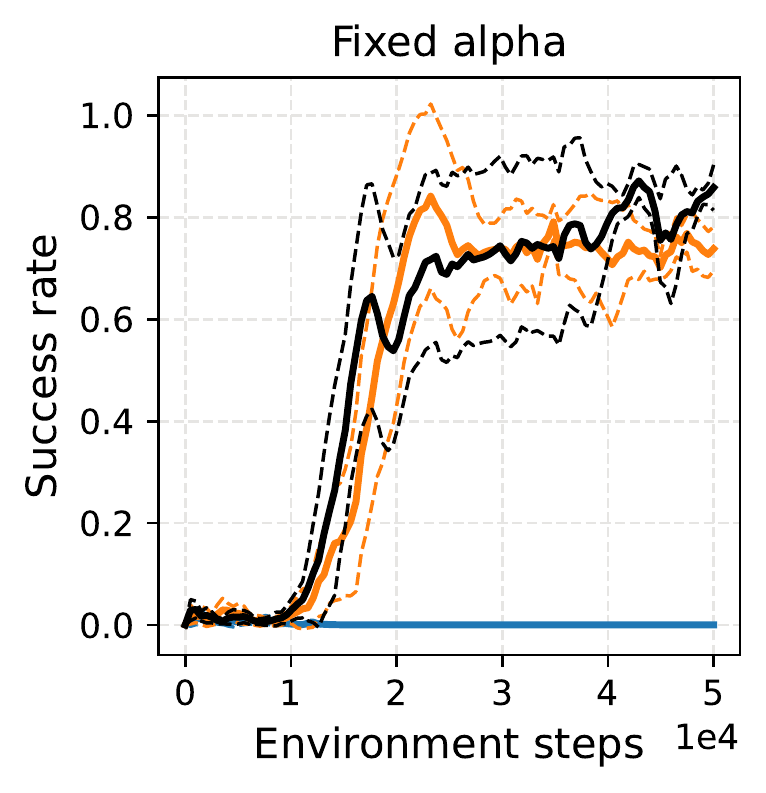}&
        \includegraphics[width=0.45\columnwidth]{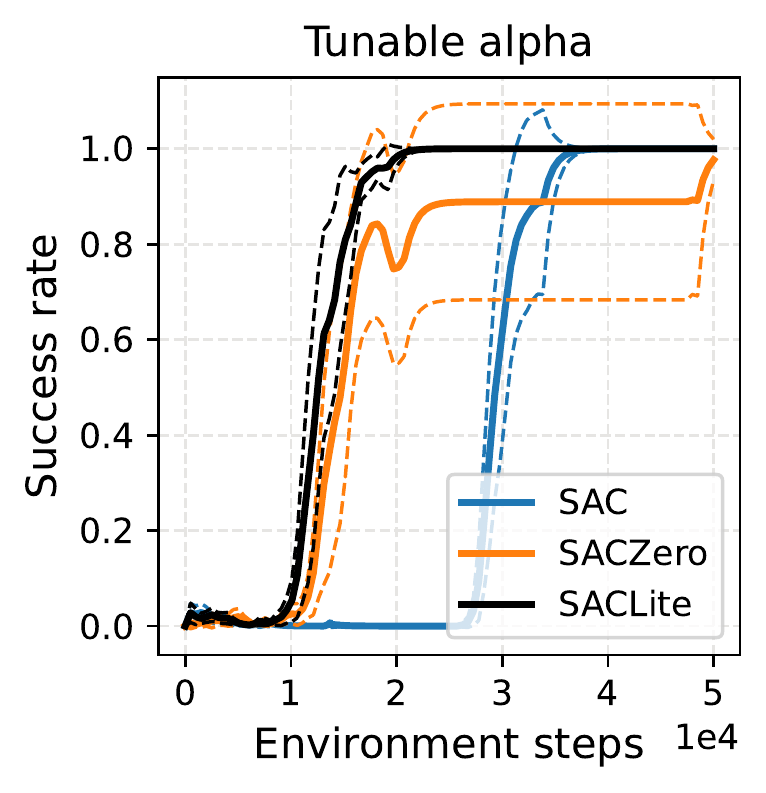}\\
        \includegraphics[width=0.44\columnwidth]{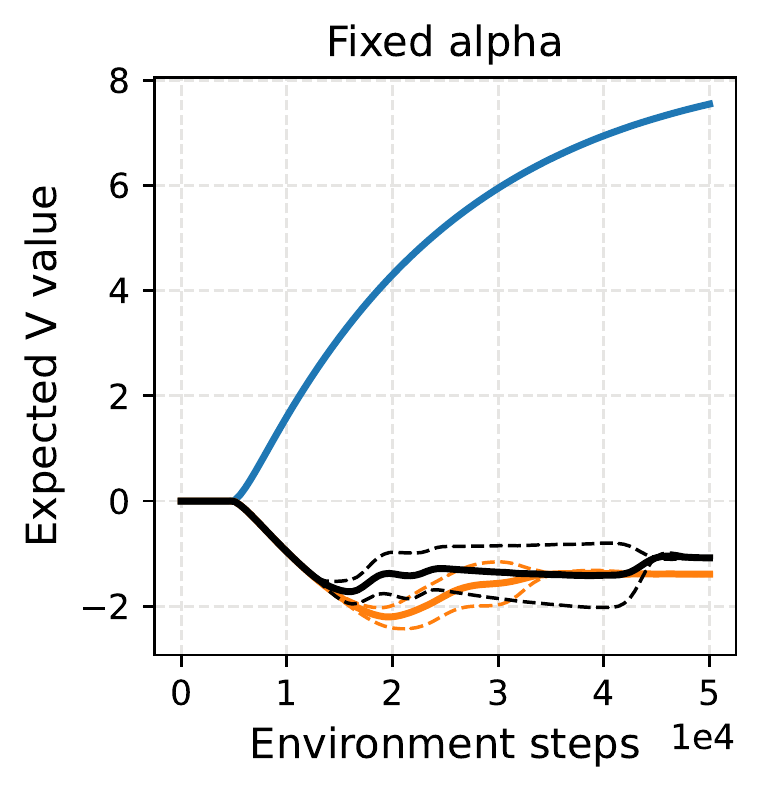}&
        \includegraphics[width=0.46\columnwidth]{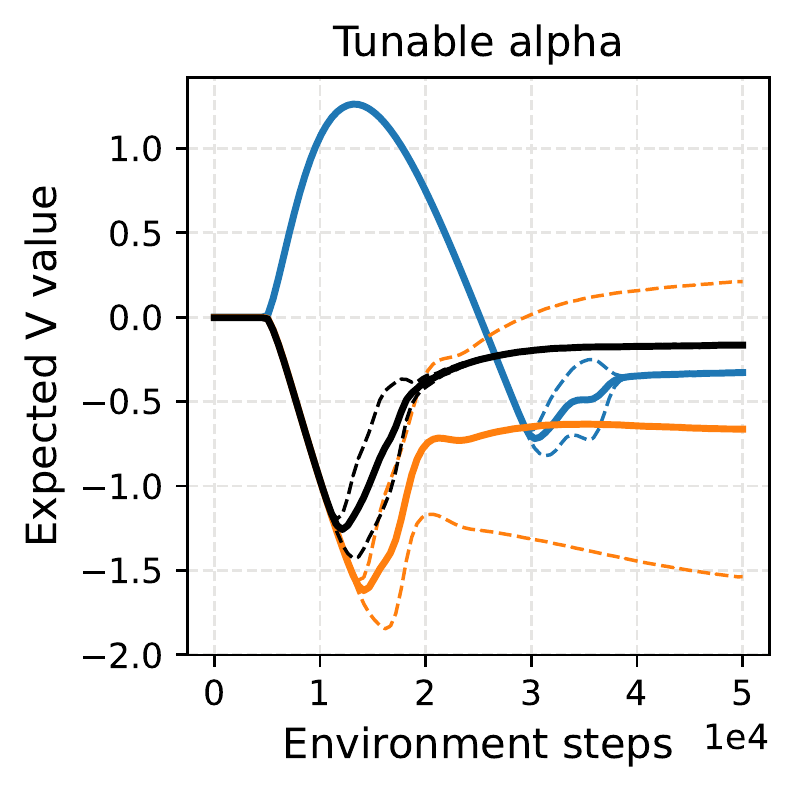}\\
    \end{tabular}
    }
    \caption{Training results on episodic \textsc{SimpleChain}.
    }
\label{fig:simple_chain_curves}
\end{center}
\vspace{-5ex}
\end{figure}

\subsection{Experiments on Episodic \textsc{SimpleChain}}
\label{sec:simple_chain_exps}
To verify our analysis in the above, we train SAC, SACLite, and SACZero on episodic \textsc{SimpleChain}, and plot their training 
success rate curves in Figure~\ref{fig:simple_chain_curves}, where a success is defined as the agent reaching the terminal state 
in 50 steps.
We also plot their expected V values ($V\triangleq\expect_{s\sim\mathcal{D},a\sim\pi}Q(s,a)$) during the training.
Note that the hyperparameters are all the \emph{same} across the three methods, and the only difference is how each deals with entropy.
Each method was run with 9 random seeds, and the dashed lines denote the upper and lower bounds of the 95\% confidence interval (CI)
for the solid curve with the same color\footnote{In the following, our training plots will all follow this convention of 
 computing a 95\% CI with 9 random seeds and using dashed lines to represent the CI, unless otherwise stated.}.
We see that the empirical results exactly match our earlier analysis: 
\begin{compactenum}[1)]
    \item With a fixed $\alpha$, SAC hardly makes any progress because the policy never wants to terminate. 
    Its V value keeps increasing during training, due to the entropy rewards.
    Both SACLite and SACZero obtain good performance, and their V values evolve reasonably, 
    more accurately reflecting the actual returns of the task reward.
    \item When $\alpha$ is tunable, even though SAC eventually obtains a perfect success rate, its sample efficiency is 
    worse. 
    Its V value is greatly bloated by the entropy rewards before dropping to the correct level.
    Generally, when specifying an initial value for $\alpha$, there is a conflict between more exploration and less \effect{}.
\end{compactenum}

\begin{figure}[t!]
\begin{center}
\resizebox{0.9\columnwidth}{!}{
    \begin{tabular}{@{}c@{}}
        \begin{tabular}{@{}cc@{}}
            \includegraphics[width=0.5\columnwidth]{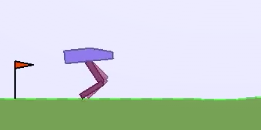}&
            \includegraphics[width=0.5\columnwidth]{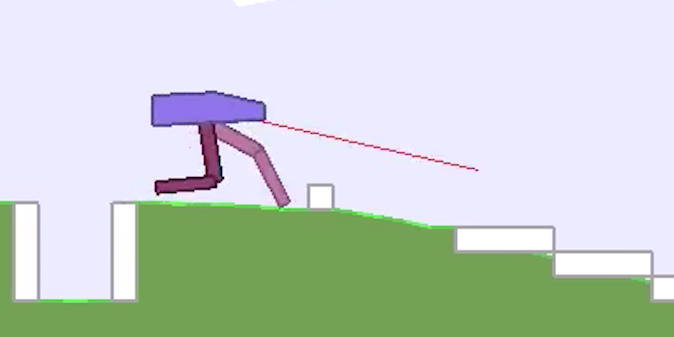}\\
        \end{tabular}
        \begin{tabular}{c@{}c@{}c@{}c@{}}
            \includegraphics[width=0.25\columnwidth]{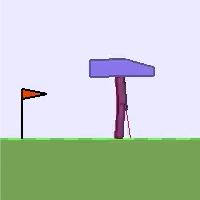}&
            \includegraphics[width=0.25\columnwidth]{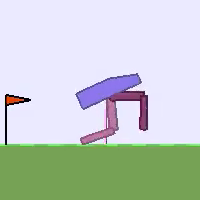}&
            \includegraphics[width=0.25\columnwidth]{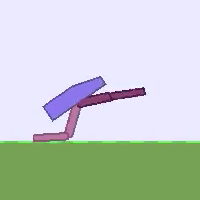}&
            \includegraphics[width=0.25\columnwidth]{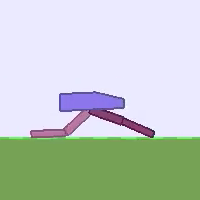}\\
        \end{tabular}
    \end{tabular}
}
\caption{Bipedal-walker tasks. Left and Middle: example frames of the normal and hard versions.
    Right: one failure case of SAC (four key frames are shown), where the robot hacks entropy bonuses by doing the splits to 
    increase episode lengths.
    }
\label{fig:bipedal_walkers}
\end{center}
\vspace{-3ex}
\end{figure}

\begin{figure}[t!]
\begin{center}
\resizebox{0.8\columnwidth}{!}{
    \begin{tabular}{@{}c@{}c@{}}
        \includegraphics[width=0.45\columnwidth]{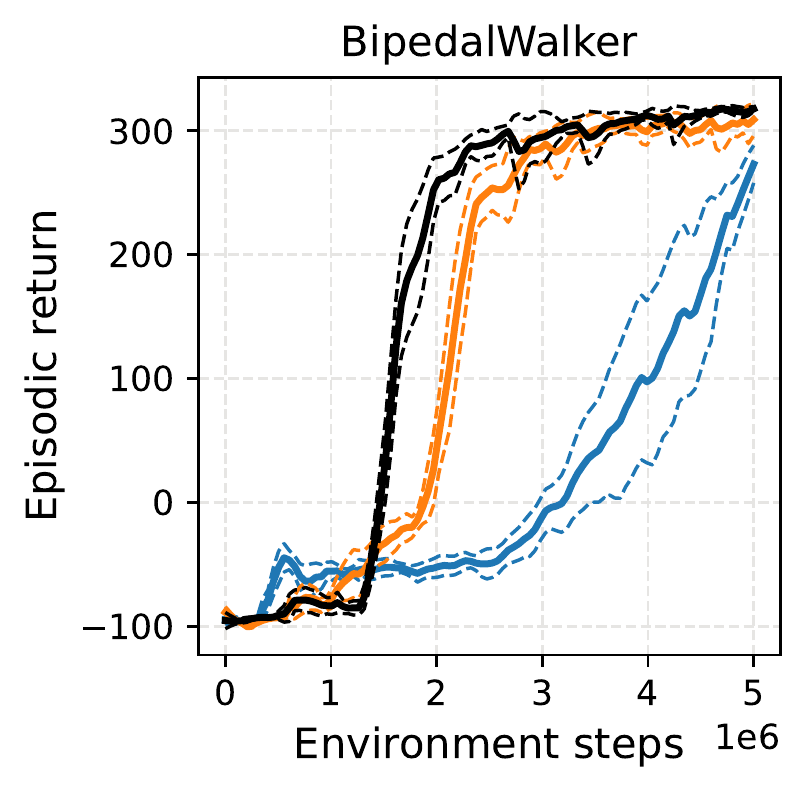}&
        \includegraphics[width=0.45\columnwidth]{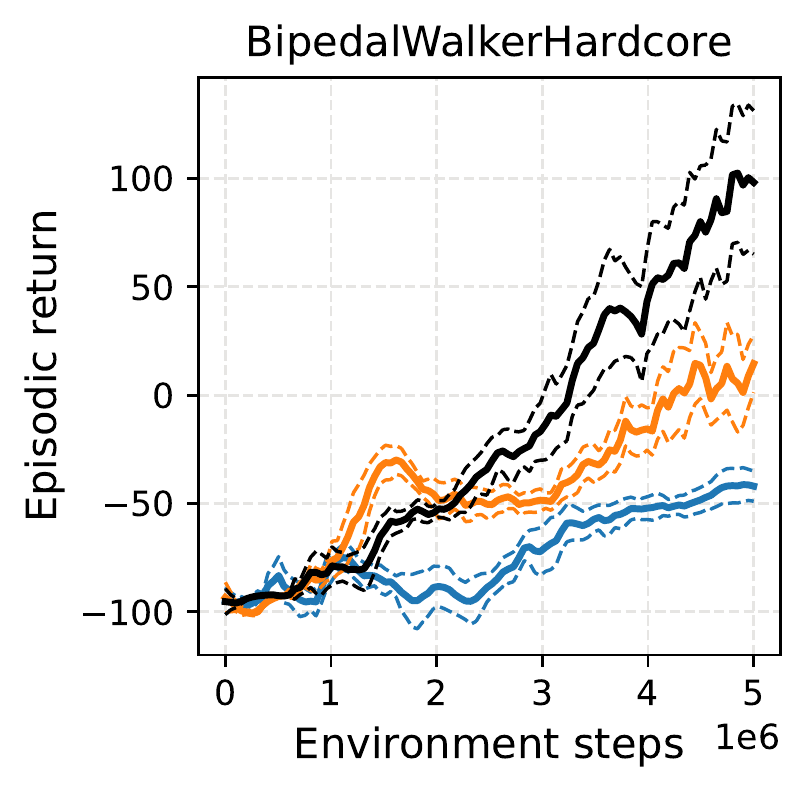}\\
        \includegraphics[width=0.45\columnwidth]{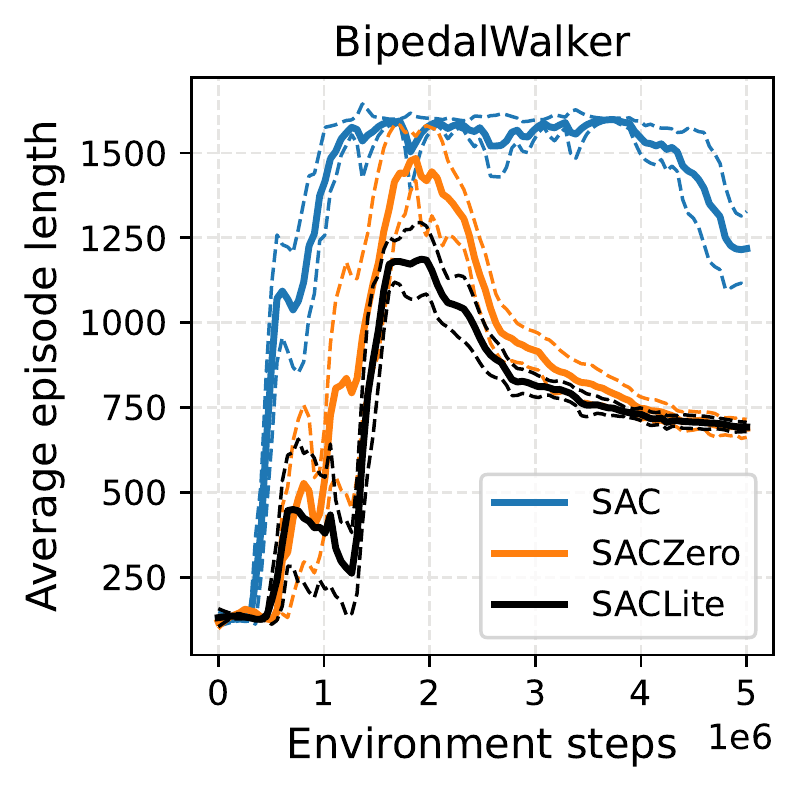}&
        \includegraphics[width=0.43\columnwidth]{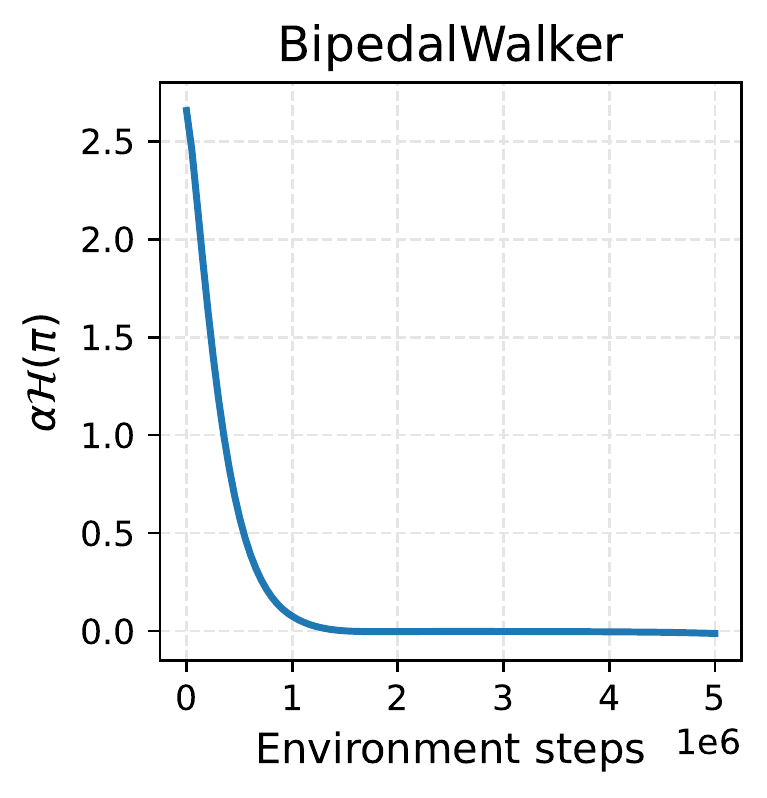}\\        
    \end{tabular}
}
\caption{Training curves of the two bipedal-walker tasks.}
\label{fig:bipedal_curves}
\end{center}
\vspace{-3ex}
\end{figure}

\subsection{Experiments on the Bipedal-walker Tasks}
\label{sec:bipedalwalker_exps}
To further illustrate the \effect{} of entropy rewards in more natural episodic settings, we test on two representative bipedal-walker tasks~\citep{Brockman2016}.
In either task (Figure~\ref{fig:bipedal_walkers}, left and middle), a bipedal robot is motivated to move forward 
until a far destination to collect a total reward of $300$. 
The reward is evenly distributed on the $x$ axis to guide the robot.
An episode ends with a reward of $-100$ if the robot falls.
%
%A small action penalty is also applied to every step.
%
The terrain is randomly generated for each episode, and the hardcore version additionally adds random obstacles, pits, stairs, etc to the terrain.
%
%The observation space is 24D and the action space is 4D.

Again we train SAC, SACLite, and SACZero with an initial $\alpha=1$ on both tasks and plot their training curves in Figure~\ref{fig:bipedal_curves}.
Within 5M environment steps, SAC's sample efficiency and final performance are greatly reduced by the \effect{} on \textsc{BipedalWalker} (compared to SACZero).
Even with normalization, the entropy reward still hurts the performance on \textsc{BipedalWalkerHardcore} (SACZero \vs\ SACLite).
Surprisingly, SACLite is already able to achieve top performance consistently even without entropy rewards.

On \textsc{BipedalWalker}, after evaluating an intermediate model of SAC at 2M steps when its performance still struggles to increase compared to SACLite and SACZero, we find that a failure case is that the robot does the splits on the ground without moving forward (Figure~\ref{fig:bipedal_walkers}, right).
This behavior started right after 1M steps, since when most episode lengths reached the time limit of $1600$ (Figure~\ref{fig:bipedal_curves}, lower left).
It lets the robot harvest positive entropy rewards before they decay at about 1.2M steps due to $\alpha$ (Figure~\ref{fig:bipedal_curves}, lower right).
Even though the entropy bonus is very small at 2M steps, it still took the policy quite some time time to recover from this behavior.
In comparison, both SACLite and SACZero were able to quickly get rid of this sub-optimal transient behavior, because they don't have any entropy reward or its \effect{} is reduced by zero-mean normalization.

\begin{figure}[t!]
\begin{center}
\resizebox{0.8\columnwidth}{!}{
    \begin{tabular}{@{}c@{}c@{}}
        \includegraphics[width=0.45\columnwidth]{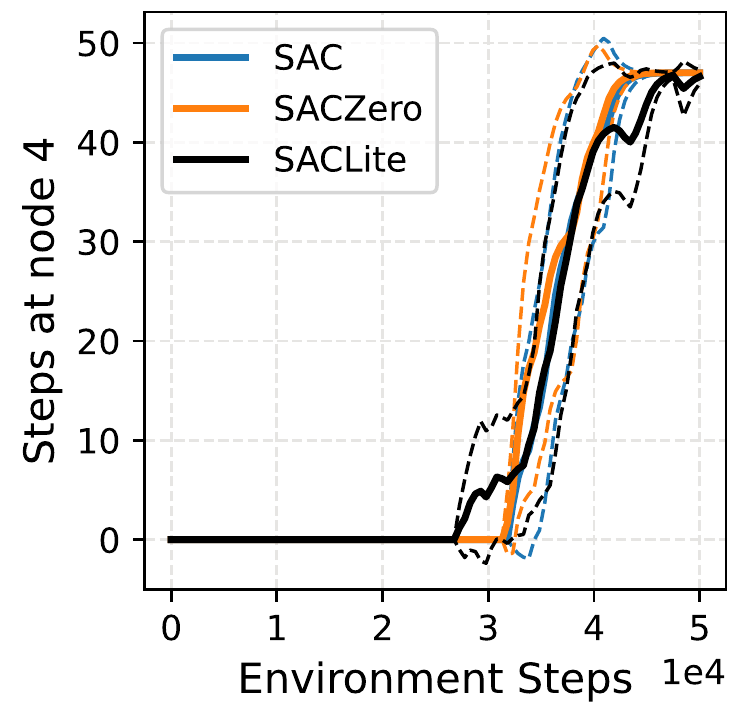}&
        \includegraphics[width=0.45\columnwidth]{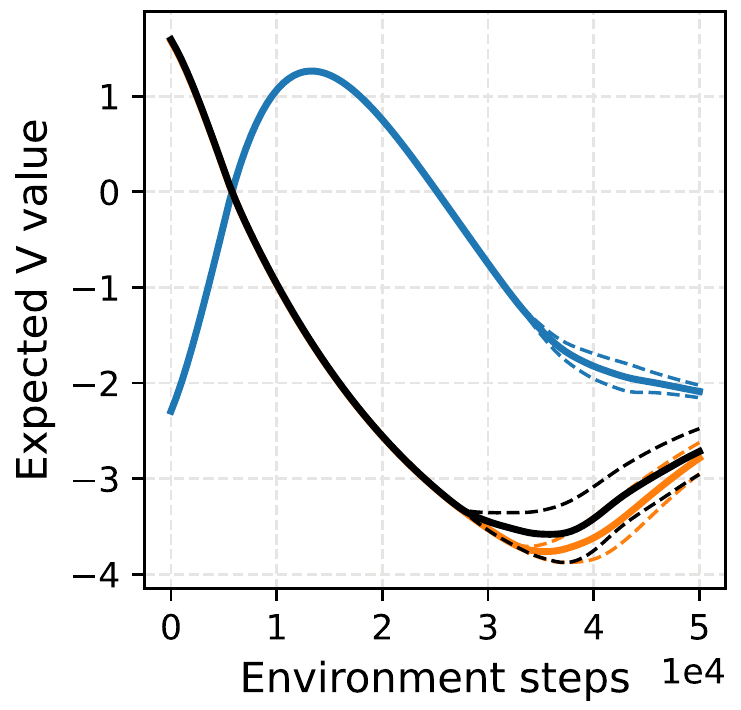}\\
    \end{tabular}
}
\caption{Training curves of \textsc{SimpleChain} with an infinite horizon.
All three methods reach the optimal 47 steps at node 4.
%
%SACZero and SACLite have V value curves more interpretable than SAC.
}
\label{fig:inf_simple_chain_curves}
\end{center}
\vspace{-6ex}
\end{figure}

\subsection{\textsc{SimpleChain} with an Infinite Horizon}
\label{sec:simple_chain_inf}
One would imagine that the \effect{} of entropy rewards will disappear in an infinite-horizon setting, because now 
there are no longer terminal states and entropy rewards will be added to all states.
Theoretically, such a reward translation makes no difference to the optimal policy in an infinite-horizon setting.
To verify this, we define an infinite-horizon version of \textsc{SimpleChain} (Figure~\ref{fig:simple_chain}, right).
It has the same MDP with the episodic version, except that node 4 is not a terminal state.
We train SAC, SACZero, and SACLite with a tunable $\alpha$ on it with an episode time limit of $50$.
Figure~\ref{fig:inf_simple_chain_curves} shows the curves of steps staying at node 4, and they do match our expectation:
all three methods obtain similar perfect performance.
Even though the expected V value of SAC still fluctuates due to the entropy reward, for this simple task the fluctuation doesn't hinder policy learning.
However, on the other hand, the entropy reward doesn't show any obvious benefit of encouraging exploration in this task (SAC/SACZero compared to SACLite).

%Combining this result with the results from Section~\ref{sec:simple_chain_exps} and Section~\ref{sec:bipedalwalker_exps}, 

So far the downside of \effect{} outweighs its advantages in episodic settings.
And in a preliminary infinite-horizon case, the entropy reward (SAC/SACZero) hasn't really been superior to the entropy cost (SACLite).
%
%Thus we ask the questions:
%
%\begin{displayquote}
    %
    Are these results due to our current test tasks being too simple or special? 
    Will the necessity of the entropy reward, including performance improvement and policy robustness, 
    be justified on many other tasks in general?
    %
%\end{displayquote}
%
%We will answer them in the remaining sections.

\section{A Large-scale Empirical Study}
%
%So far, there have been few empirical investigations into whether the entropy cost or the entropy reward plays a more important role in 
%encouraging exploration for SAC, or if they are equally important.
%
In this section we conduct a large scale of empirical study, covering various scenarios of continuous control.
Again we will compare SAC, SACZero, and SACLite as introduced in Section~\ref{sec:leveling_effect} with 
an initial entropy weight $\alpha=1$ to be tuned automatically.
For some tasks, we also experiment with an initial $\alpha=0.1$ and name the corresponding methods as SAC-a01, SACZero-a01, and SACLite-a01.
All other hyperparameters are fixed given a task.

\textbf{A brief summary of our findings}\ \ The entropy cost plays a major, if not entire, role of contributing to good exploration and performance. 
The entropy reward is not as necessary as it is thought to be.
Below we support this claim with various experiment results.

\begin{figure}[t!]
\begin{center}
\resizebox{\columnwidth}{!}{
\begin{tabular}{@{}c@{}}
    \begin{tabular}{@{}c@{}c@{}c@{}c@{}}
        \includegraphics[width=0.25\columnwidth]{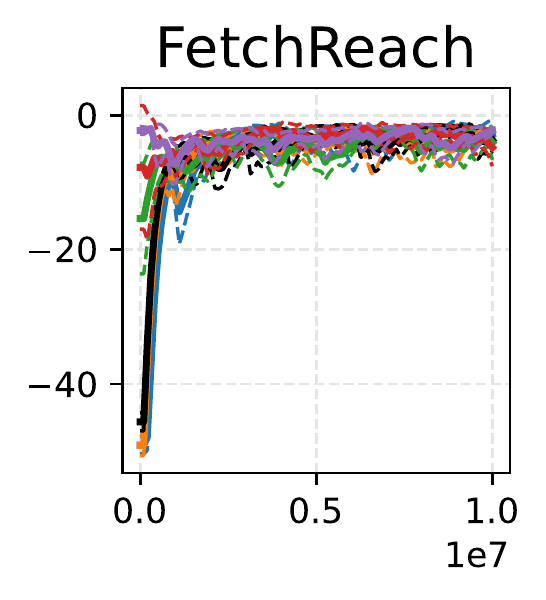}&
        \includegraphics[width=0.25\columnwidth]{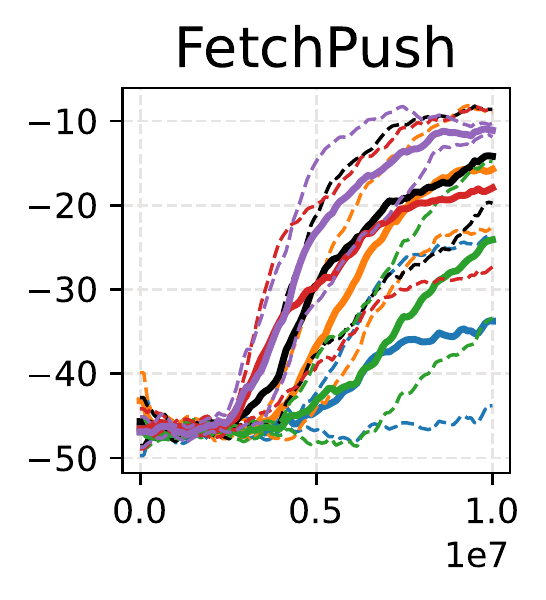}&
        \includegraphics[width=0.25\columnwidth]{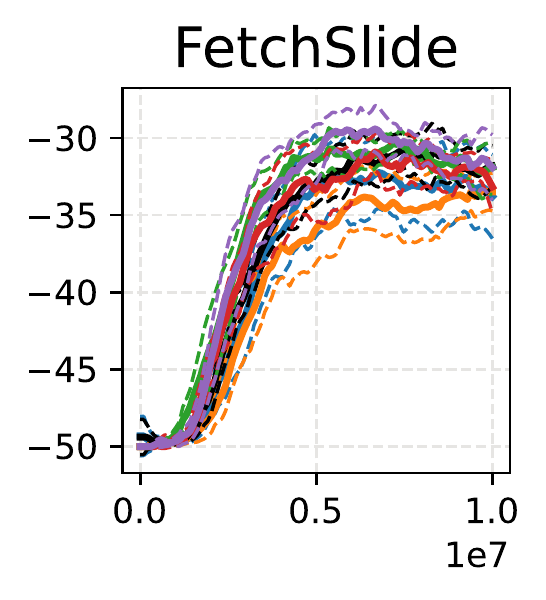}&
        \includegraphics[width=0.25\columnwidth]{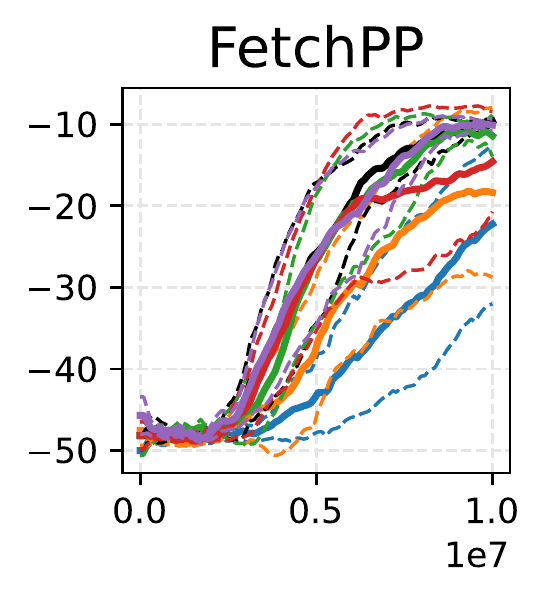}\\
    \end{tabular}\\
    \includegraphics[width=\columnwidth]{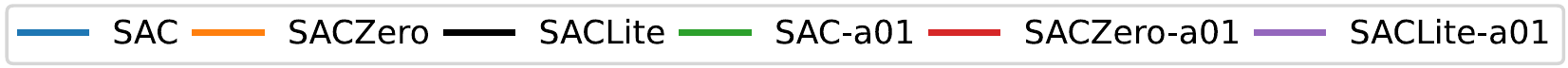}\\
\end{tabular}
}
\vspace{-3ex}
\caption{Training curves of four MuJoCo manipulation tasks. 
``FetchPP'' stands for \textsc{FetchPickAndPlace}.
$x$ axis: training environment steps; $y$ axis: episodic return.
}
\label{fig:fetch_curves}
\end{center}
\vspace{-4ex}
\end{figure}

\subsection{Single-objective RL}
\label{sec:manipulation}
\textbf{Manipulation}\ \ We start with four challenging MuJoCo manipulation tasks ~\citep{plappert2018multigoal} that have infinite horizons and sparse rewards: \textsc{FetchReach},
\textsc{FetchPush}, \textsc{FetchSlide}, and \textsc{FetchPickAndPlace}.
In each task, a robot arm is required to accomplish a goal of reaching, pushing a box, sliding a puck, or picking up a box to a desired location which will be randomly sampled.
The reward is $0$ when a goal is achieved and $-1$ otherwise.
The time limit of each episode is $50$ steps.
We train each task to 10M environment steps  with the same hyperparameters suggested by \citet{plappert2018multigoal}.

According to our analysis in Section~\ref{sec:simple_chain_inf}, in theory the \effect{} of the entropy reward no longer exists in an infinite-horizon setting.
However, Figure~\ref{fig:fetch_curves} shows that on \textsc{FetchPush} and \textsc{FetchPickAndPlace}, SAC is much worse than SACLite, clearly affected by 
the entropy reward.
There are two hypotheses for this observation. 
\begin{compactenum}[(a)]
\iffalse
    \item We use \emph{biased} target Q values for bootstrapping in TD backup. 
    %
    When target Q networks are initialized to output (near-)zero values, initially there will still be \effect{}.
    %
    Infrequent states (including those that are successful) are less trained in TD backup, and the functionally approximated values of these states are more biased towards zeros.
    %
    Thus with the entropy bonus, the policy tends to visit normal states due to their higher values.
\fi
    %
    \item The overall mean of the entropy reward is constantly changing because of matching to the entropy target, and this adds another layer of non-stationarity to the value learning dynamics, even though we can assume that at any moment this mean entropy reward is benign.
    \item The entropy reward, as an intrinsic reward, indeed unfortunately obscures the task reward, making the policy learn sub-optimal or task-unrelated behaviors.
\end{compactenum}

We additionally train SAC-a01, SACZero-a01, and SACLite-a01 and add their curves to Figure~\ref{fig:fetch_curves}.
On \textsc{FetchPush}, (a) seems primary because SACZero is comparable to SACLite (with zero-mean normalization (a) is no longer present). 
However, (b) cannot be completely ruled out because on \textsc{FetchSlide} and \textsc{FetchPickAndPlace}, SACZero is clearly worse than SACLite. 
With a smaller initial $\alpha$, SACZero-a01 is closer to SACLite and SACLite-a01, indicating that the performance gets improved if the zero-mean entropy reward is less weighted from the beginning. 
Both hypotheses seem to account for the results on \textsc{FetchPickAndPlace}, given that SACLite$>$SACZero$>$SAC. 
Note that a smaller initial $\alpha$ will not eradicate the issues faced by SAC, as shown by \textsc{FetchPush}.
%
%In a word, on this set of tasks, all the three reasons seem to account for the observed results.

\begin{figure}[t!]
\begin{center}
\resizebox{\columnwidth}{!}{
\begin{tabular}{@{}c@{}}
    \begin{tabular}{@{}c@{}c@{}c@{}c@{}}
        \includegraphics[width=0.25\columnwidth]{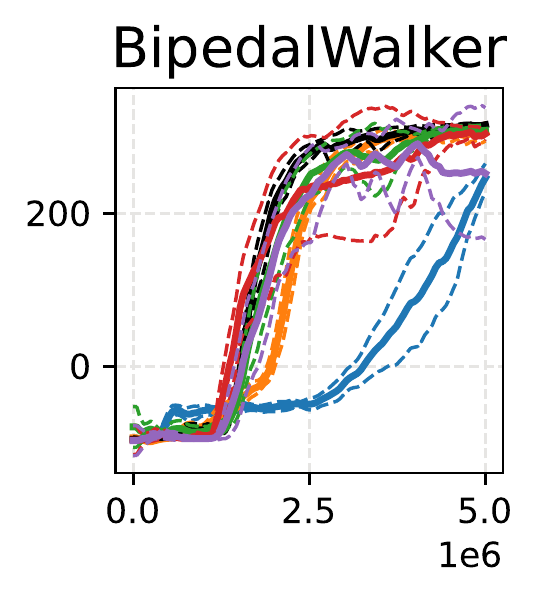}&
        \includegraphics[width=0.26\columnwidth]{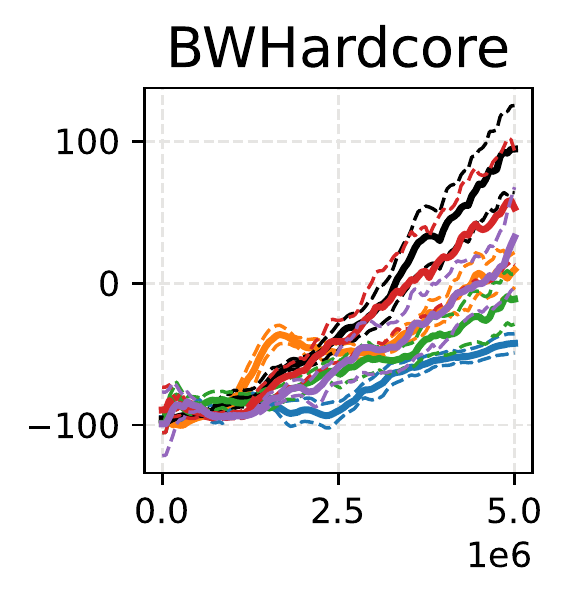}&
        \includegraphics[width=0.26\columnwidth]{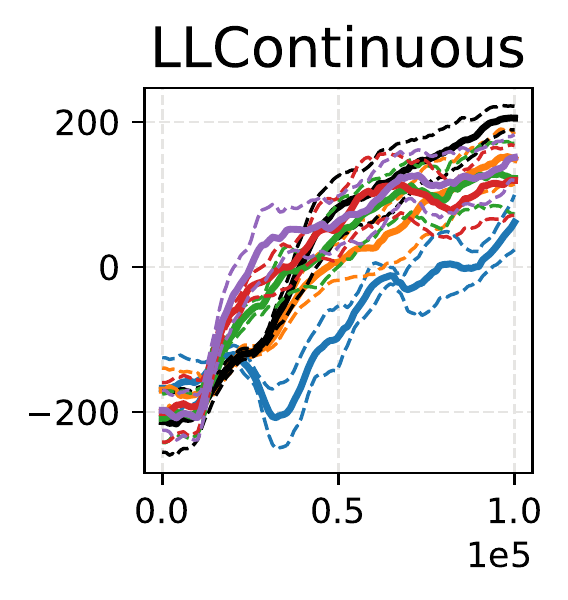}&
        \includegraphics[width=0.25\columnwidth]{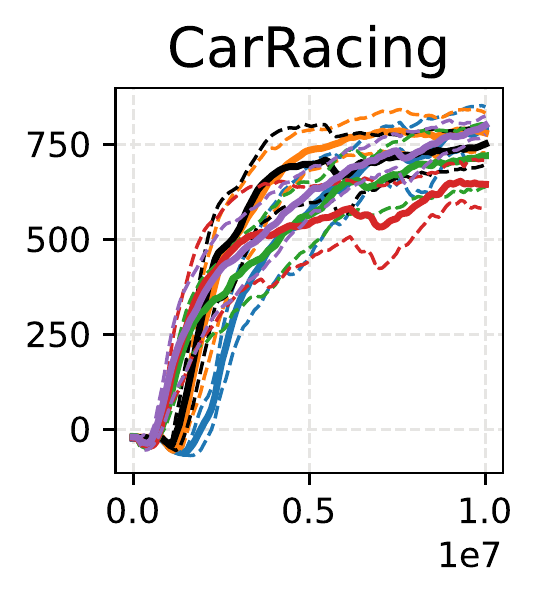}\\
    \end{tabular}\\
    \includegraphics[width=\columnwidth]{images/legend.pdf}\\
\end{tabular}
}
\vspace{-3ex}
\caption{Training curves of four Box2D tasks. 
``BWHardcore'' stands for \textsc{BipedalWalkerHardcore} and ``LLContinuous'' stands for \textsc{LunarLanderContinuous}.
$x$ axis: training environment steps; $y$ axis: episodic return.
}
\label{fig:box2d_curves}
\end{center}
\vspace{-4ex}
\end{figure}

\textbf{Box2D}\ \ The next four Box2D control tasks \citep{Brockman2016} to be evaluated are episodic.
\textsc{BipedalWalker} and~\textsc{BipedalWalkerHardcore} have been partially evaluated in Section~\ref{sec:bipedalwalker_exps}, but here we additionally test 
SAC-a01, SACZero-a01, and SACLite-a01 on them for further analysis. 
The other two tasks are \textsc{LunarLanderContinuous} and \textsc{CarRacing}.
In \textsc{LunarLanderContinuous}, a lander is rewarded for resting on a landing pad and penalized for crashing during the landing process.
In \textsc{CarRacing}, a racing car is rewarded for each tile visited on the track and penalized for being too far away from the track.
Both tasks have an episode time limit of $1000$ steps.
Figure~\ref{fig:box2d_curves} shows that while sometimes an initial $\alpha=0.1$ helps alleviate the side effects of the entropy reward in SAC (\textsc{BipedalWalker} and \textsc{LunarLanderContinuous}), it can hurt the performance of SACLite (\textsc{BipedalWalkerHardcore}) and SACZero (\textsc{CarRacing}) if they obtain good results with $\alpha=1$. 
In other words, to achieve top results, sufficient exploration by a large $\alpha$ is necessary for some tasks.
Note that SACLite is able to perform well consistently with $\alpha=1$.

\begin{figure}[t!]
\begin{center}
\resizebox{\columnwidth}{!}{
    \begin{tabular}{@{}c@{}c@{}c@{}c@{}}
        \includegraphics[width=0.25\columnwidth]{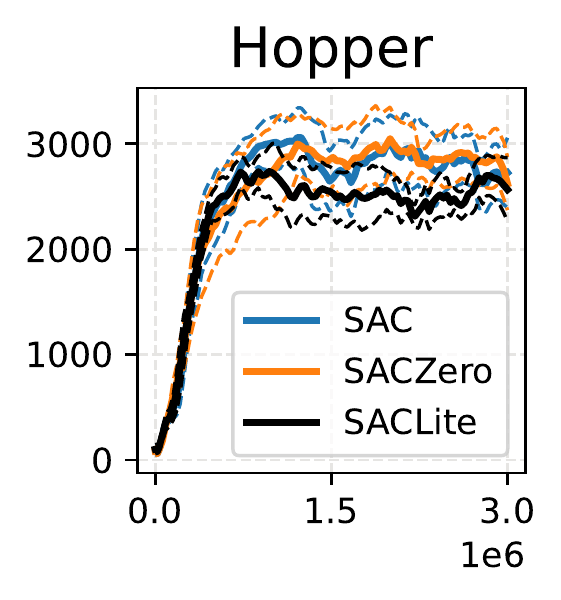}&
        \includegraphics[width=0.25\columnwidth]{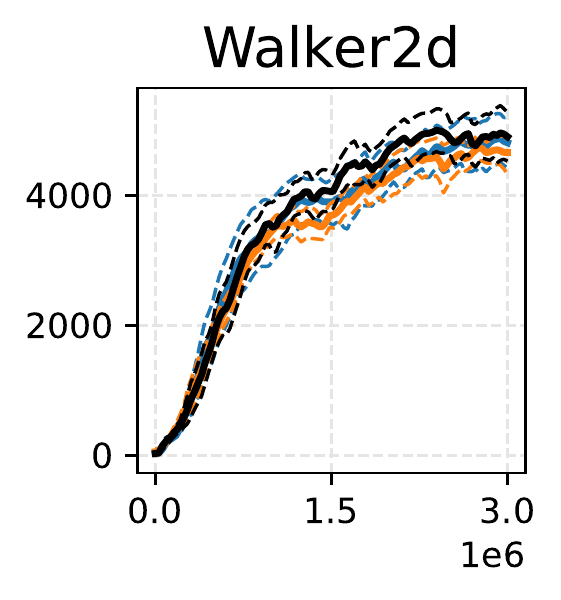}&
        \includegraphics[width=0.25\columnwidth]{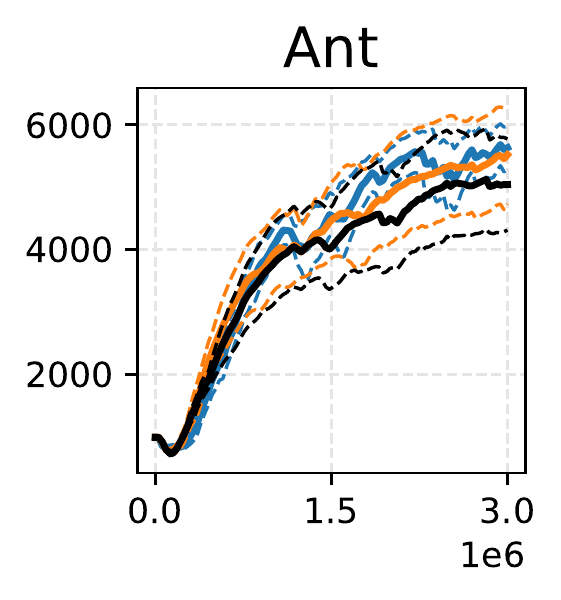}&
        \includegraphics[width=0.26\columnwidth]{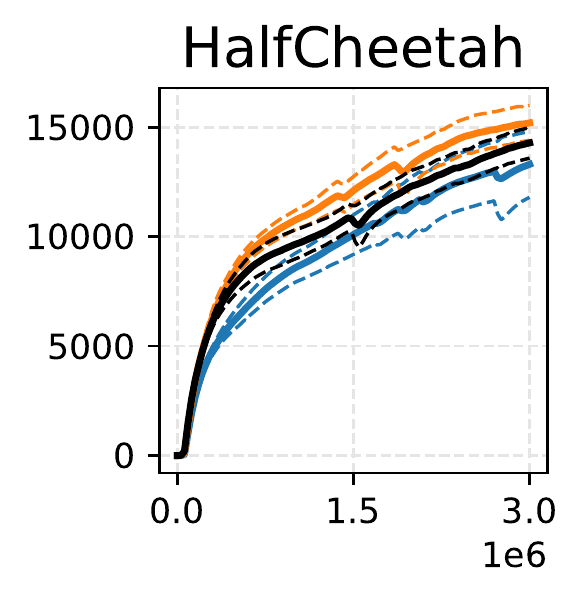}\\
    \end{tabular}
}
\vspace{-3ex}
\caption{Training curves of four locomotion tasks.
$x$ axis: training environment steps; $y$ axis: episodic return.
}
\label{fig:locomotion_curves}
\end{center}
\vspace{-3ex}
\end{figure}

\begin{figure}[t!]
\begin{center}
\resizebox{\columnwidth}{!}{
    \begin{tabular}{@{}c@{}c@{}c@{}c@{}}
        \includegraphics[width=0.25\columnwidth]{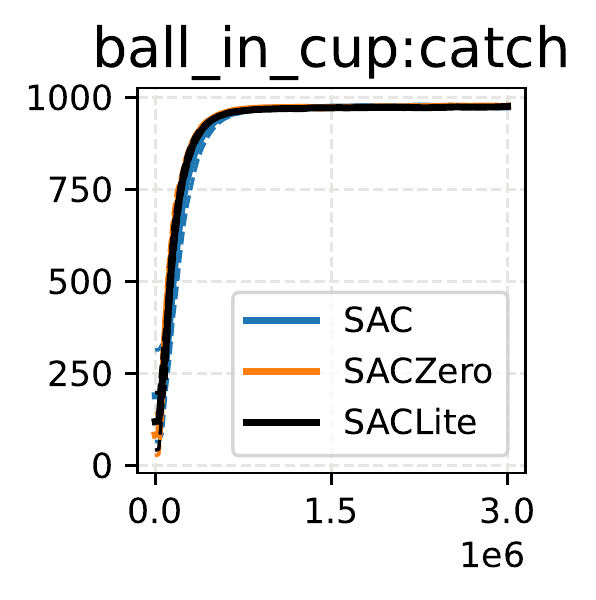}&
        \includegraphics[width=0.25\columnwidth]{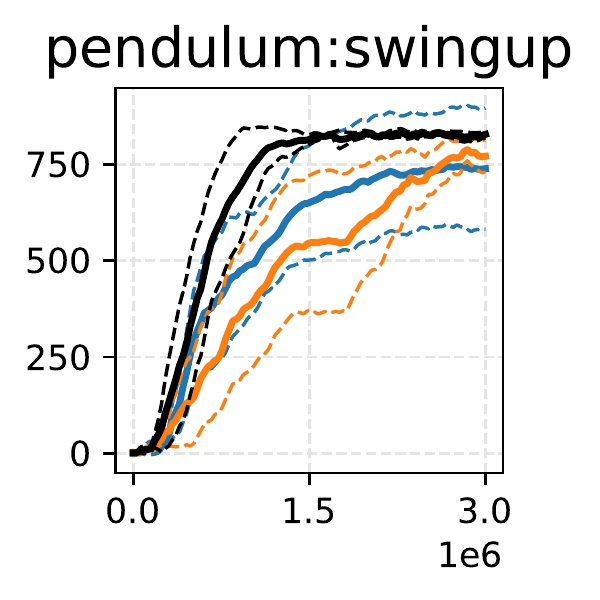}&
        \includegraphics[width=0.24\columnwidth]{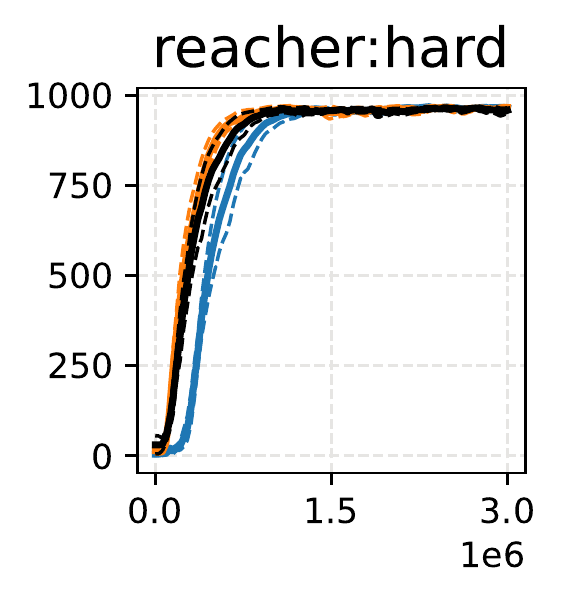}&
        \includegraphics[width=0.25\columnwidth]{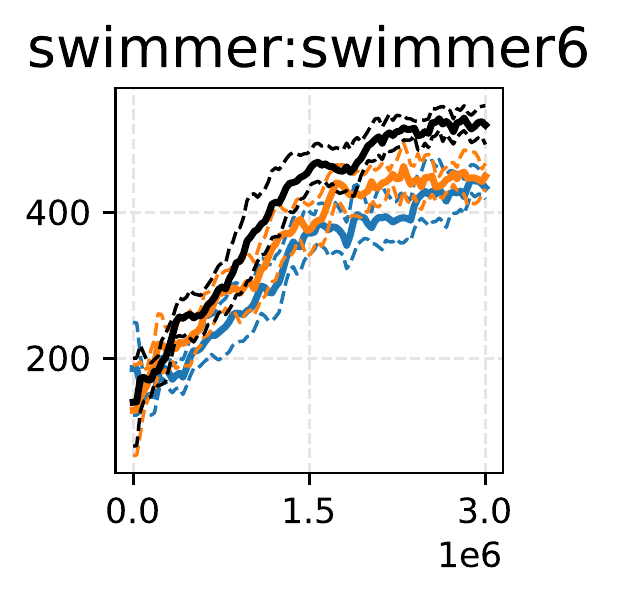}\\
        \includegraphics[width=0.25\columnwidth]{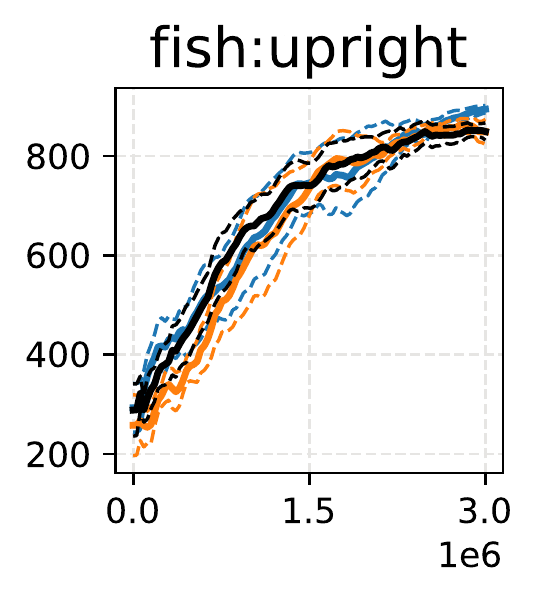}&
        \includegraphics[width=0.25\columnwidth]{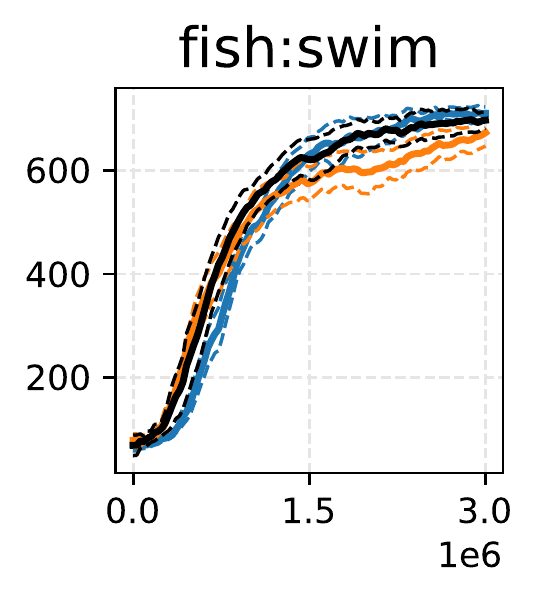}&
        \includegraphics[width=0.25\columnwidth]{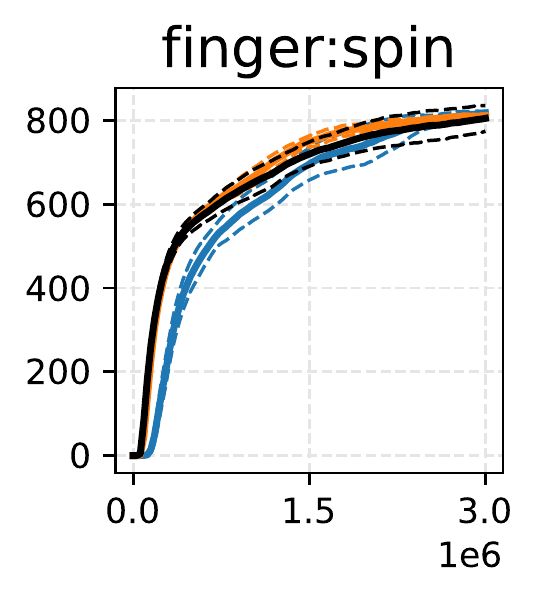}&
        \includegraphics[width=0.25\columnwidth]{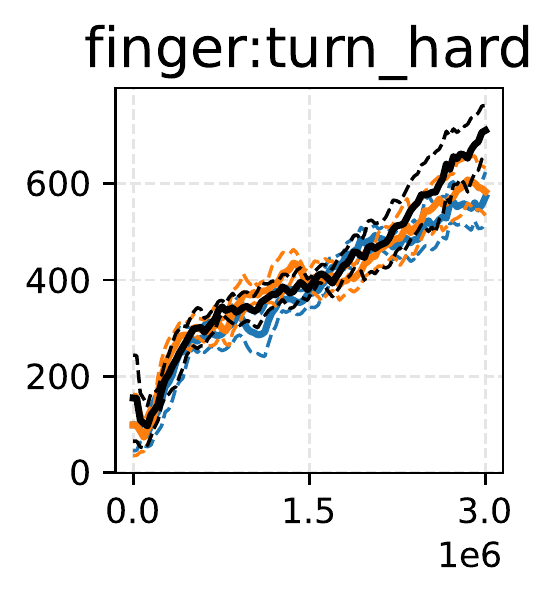}\\
    \end{tabular}
}
\vspace{-3ex}
\caption{Training curves of eight DM control tasks.
$x$ axis: training environment steps; $y$ axis: episodic return.
}
\label{fig:dm_control_curves}
\end{center}
\vspace{-2ex}
\end{figure}

\label{sec:locomotion}
\textbf{Locomotion}\ \ We study four MuJoCo locomotion tasks that were evaluated by SAC~\citep{Haarnoja2018}: \textsc{Hopper}, \textsc{Ant}, \textsc{Walker2d}, and \textsc{HalfCheetah}.
Generally, each task rewards the robot for moving forward on the $x$ axis as fast as possible.
All tasks except \textsc{HalfCheetah} have an early termination condition when the robot is considered no longer ``healthy'', and thus are episodic.
The healthiness of a robot is determined by checking its body/joint positions.
The time limit of each episode is $1000$ steps.
We train each task with the same hyperparameters suggested by \citet{Haarnoja2018}.
Unlike the manipulation and Box2D tasks, Figure~\ref{fig:locomotion_curves} presents very close curves of SAC, SACZero, and SACLite. 
SACZero is better than SAC only on \textsc{HalfCheetah}, while SACLite doesn't have an obvious overall advantage over SAC or vice versa.

\begin{figure}[t!]
\begin{center}
\resizebox{\columnwidth}{!}{
\begin{tabular}{@{}c@{}c@{}!{\color{gray}\vrule}@{}c@{}c@{}}
    \includegraphics[width=0.25\columnwidth]{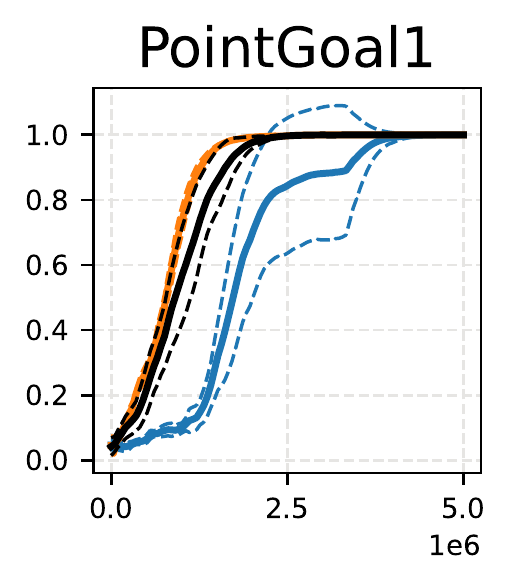}&
    \includegraphics[width=0.25\columnwidth]{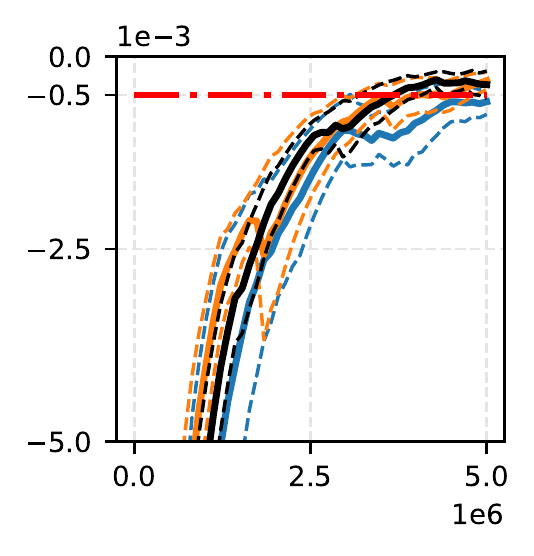}&
    \includegraphics[width=0.25\columnwidth]{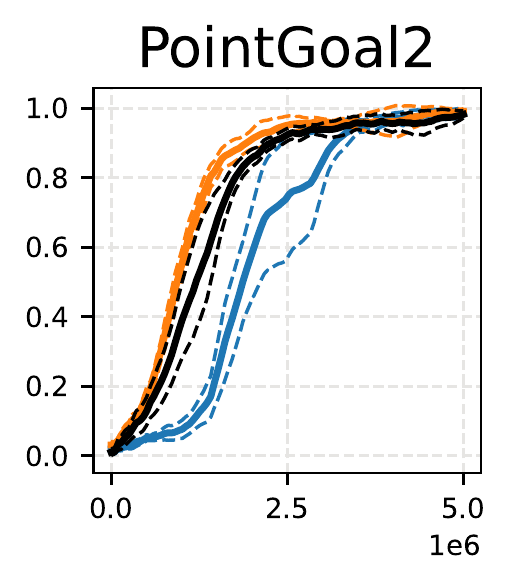}&
    \includegraphics[width=0.25\columnwidth]{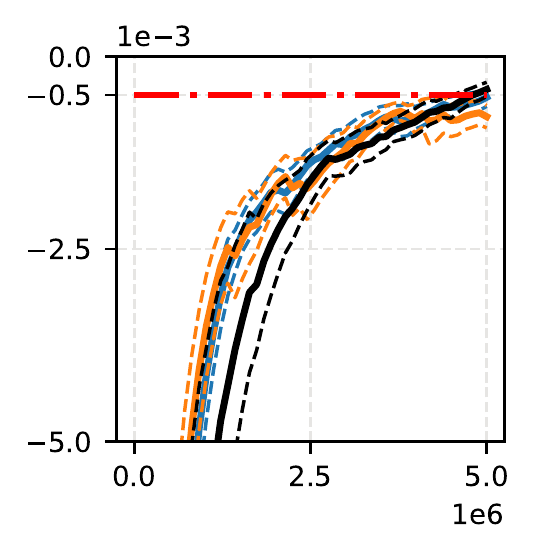}\\
    \includegraphics[width=0.25\columnwidth]{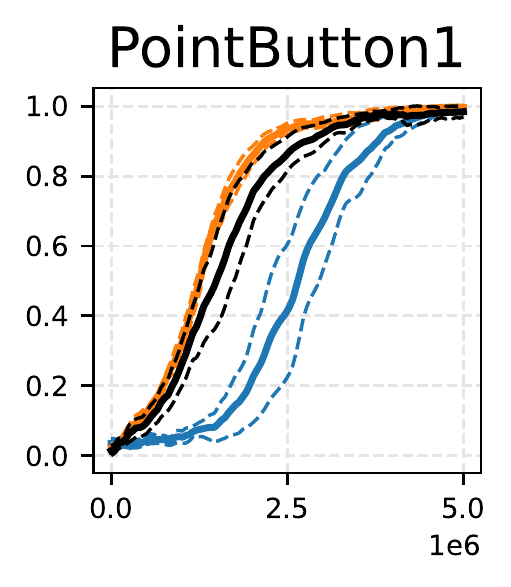}&
    \includegraphics[width=0.25\columnwidth]{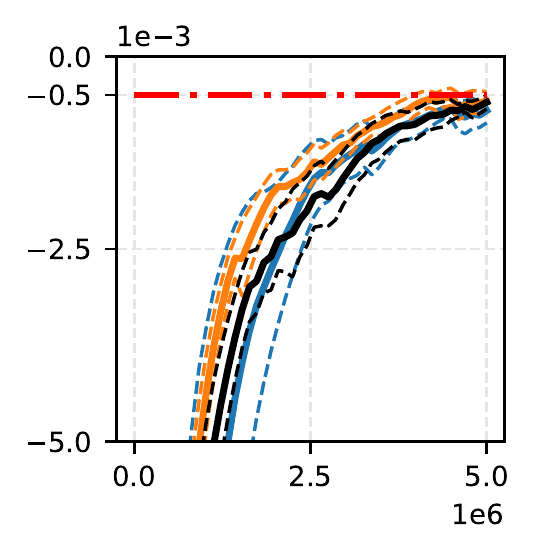}&
    \includegraphics[width=0.25\columnwidth]{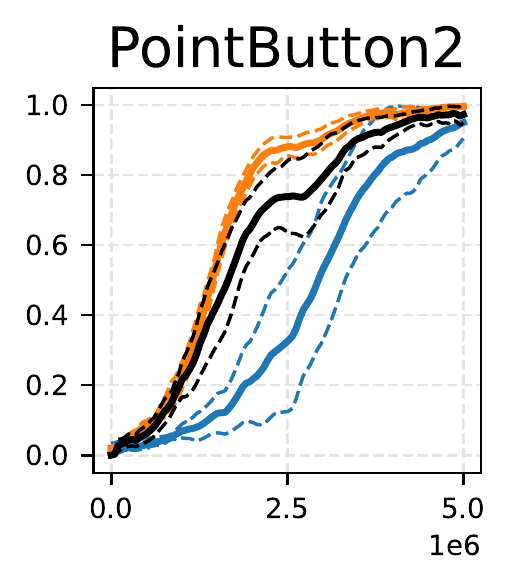}&
    \includegraphics[width=0.25\columnwidth]{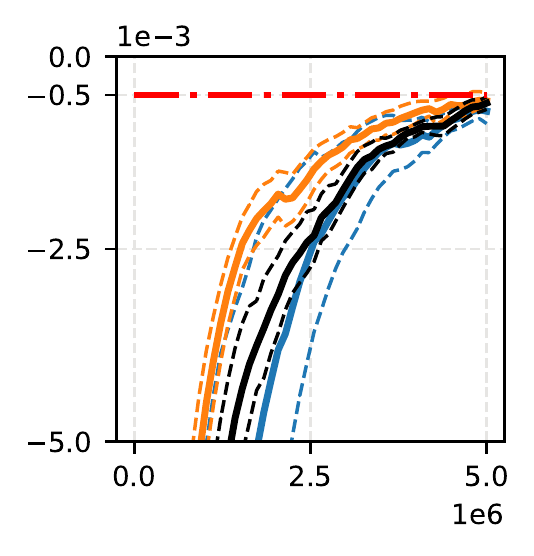}\\
    \includegraphics[width=0.25\columnwidth]{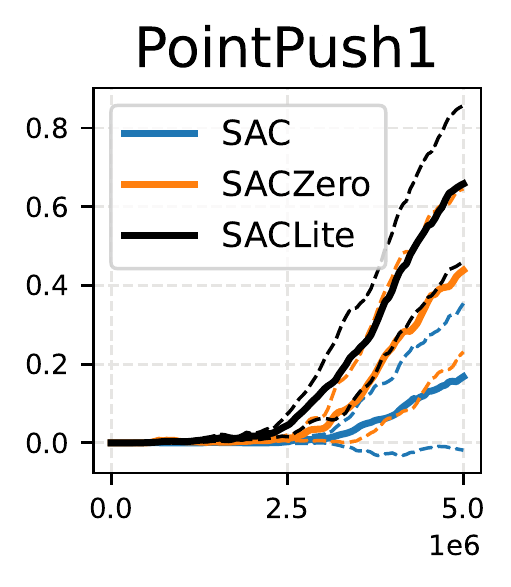}&
    \includegraphics[width=0.25\columnwidth]{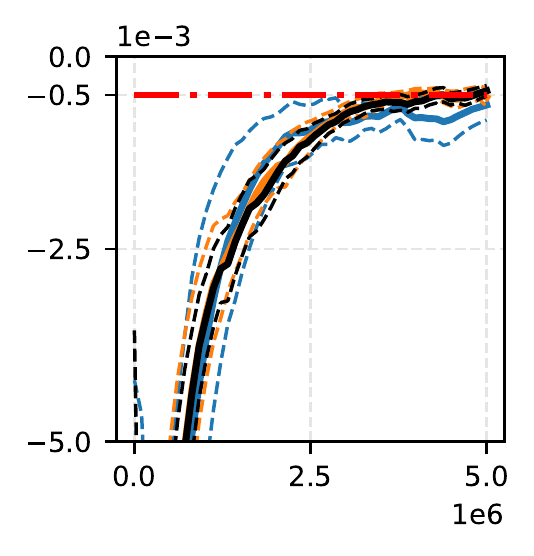}&
    \includegraphics[width=0.25\columnwidth]{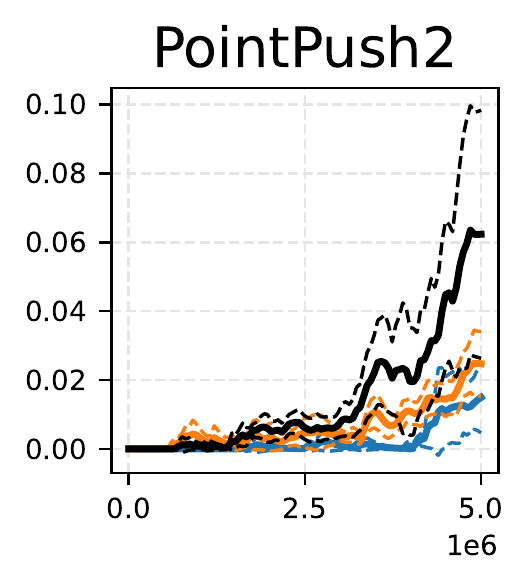}&
    \includegraphics[width=0.25\columnwidth]{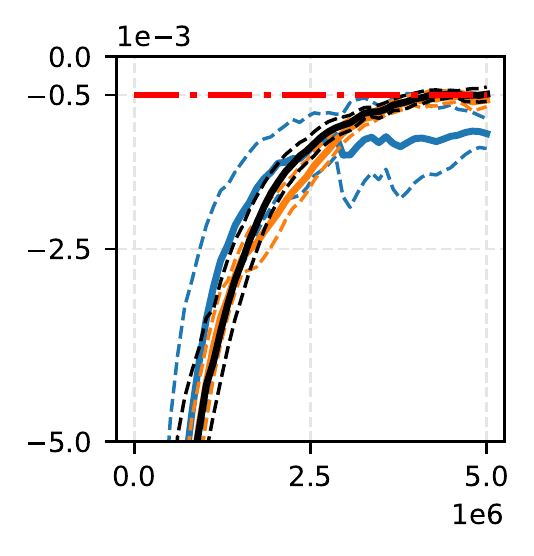}\\
\end{tabular}
}
\vspace{-2ex}
\caption{Training curves of the Safety Gym \textsc{Point} tasks. 
Odd columns are the goal success rate curves. 
Even columns are the averaged per-step constraint curves, where the red dashed line represents the constraint threshold of $5\times 10^{-4}$. 
On all tasks, SACLite and SACZero are both able to satisfy the threshold approaching the end of training.
SAC has some difficulty in constraint satisfaction on \textsc{PointPush1/2}.
$x$ axis represents training environment steps.
}
\label{fig:point_curves}
\end{center}
\vspace{-2ex}
\end{figure}

\begin{figure}[t!]
\begin{center}
\resizebox{\columnwidth}{!}{
\begin{tabular}{@{}c@{}}
    \begin{tabular}{@{}c@{}c@{}c@{}c@{}}
        \includegraphics[width=0.25\columnwidth]{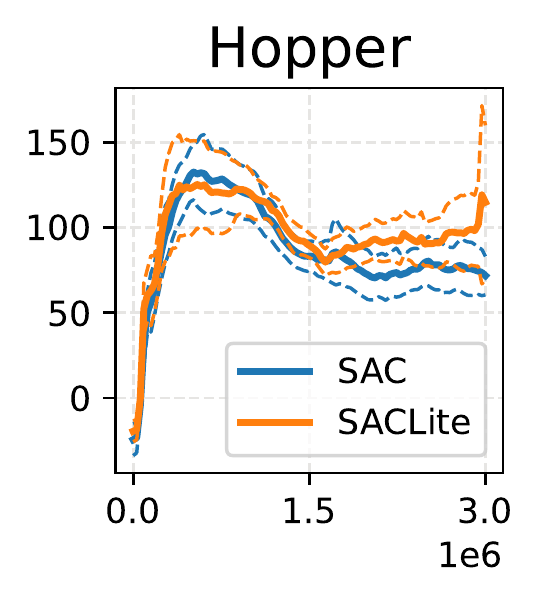}&
        \includegraphics[width=0.25\columnwidth]{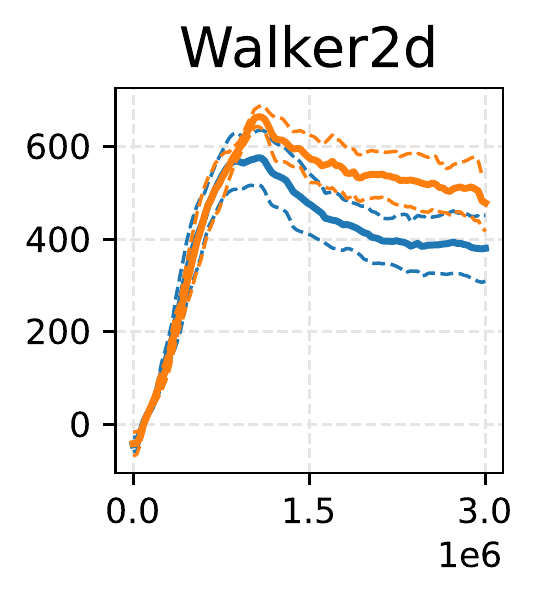}&
        \includegraphics[width=0.25\columnwidth]{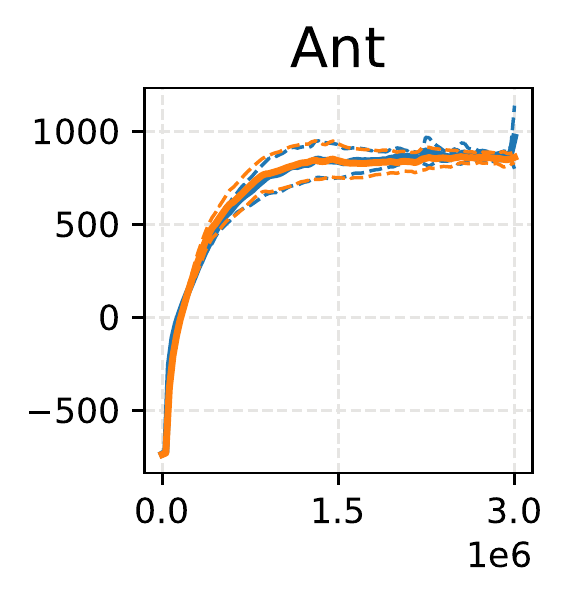}&
        \includegraphics[width=0.25\columnwidth]{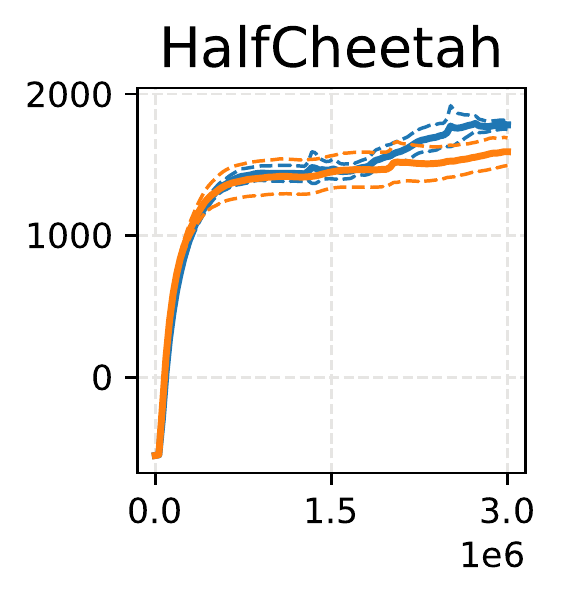}\\
    \end{tabular}\\
\end{tabular}
}
\vspace{-2ex}
\caption{Training curves of the reward robustness experiments.
$x$ axis: training environment steps; $y$ axis: gap (the smaller the better) between the evaluated returns of task and adversarial rewards.
}
\label{fig:reward_robust_curves}
\end{center}
\vspace{-2ex}
\end{figure}

%\subsubsection{DM control suite}
%
We also consider DeepMind control suite~\citep{tassa2020dmcontrol} and choose eight tasks that are in different domains with the MuJoCo ones.
All tasks require the robot to achieve some (randomized) goal state configuration starting from a distribution of initial states, for example, moving the end effector of a robot arm to a location.
A reward of $+1$ is given only when the current state satisfies the goal condition, or for some tasks the reward decreases smoothly from 1 to 0 as the state deviates more from the goal state.
The eight tasks are all infinite-horizon with a time limit of $1000$.
We train each task to 3M environment steps with every step repeating an action twice.
Figure~\ref{fig:dm_control_curves} shows that SACLite is much better than SAC on \textsc{pendulum:swingup}, \textsc{swimmer:swimmer6} and \textsc{finger:turn\_hard}, and almost the same on the other tasks.
Zero-mean normalization does not result in better performance (SACZero \vs\ SAC), indicating that the hypothesis (b) is the cause.

\subsection{Multi-objective RL}
The next scenario is multi-objective RL.
Specifically, we'd like to know what impact the entropy reward will have on the performance when multiple task rewards are already competing with each other.
Particularly, we choose a safe RL scenario where the agent needs to improve its utility while satisfying a pre-defined constraint (in expectation).
We test on six customized Safety Gym ~\citep{Ray2019} tasks of \textsc{Point}. 
Each task has an infinite horizon with an episode time limit of $1000$.
In each episode, the robot is required to achieve a random goal while avoiding obstacles, distraction objects, and/or moving gremlins. 
An episode will \emph{not} terminate when a constraint is violated or when a goal is achieved.
We set the constraint threshold to be $-5\times 10^{-4}$ per step, meaning that the agent is allowed to violate any constraint only once every 2k steps or 2 episodes in expectation. 
Note that since the agent does not have any prior knowledge about safety (it has to learn the concept from scratch), constraint satisfaction can only be asymptotic.
A success is defined as achieving the goal before timeout. 

Given the constraint threshold, we employ the Lagrangian method~\citep{Ray2019} to automatically adjust the weight $\lambda$ of the constraint reward.
When computing the loss for $\lambda$, instead of comparing the constraint V values to the accumulated discounted violation budget, we directly compare the violation rate of a rollout sample batch with the per-step threshold following
\ifdefined\isaccepted
\citet{Yu2022seditor}.
\else
\citet{Yu2022seditorAnonymous}.
\fi
We train SAC, SACZero, and SACLite for 5M steps on each task.

The experiment results are in Figure~\ref{fig:point_curves}.
On the Goal and Button tasks, SAC's sample efficiency is worse than SACLite and SACZero, although its final success rates are comparable. 
On the Push tasks, SAC struggles to improve the success rates within 5M steps and still does not satisfy the constraint threshold at the end.
At the same time, SACZero is comparable to SACLite on average.
Thus compared to the manipulation results (Section~\ref{sec:manipulation}), SAC's worse results seem to solely result from the hypothesis (a).
This is reasonable given that with the Lagrangian method adjusting the two task reward weights and causing non-stationarity in the Q values already, the extra non-stationarity by the entropy reward will further challenge the training.

\iffalse
\subsection{Locomotion with a fixed $\alpha$}
%
\label{sec:fixed_alpha}
So far all our experiments use a dynamic entropy weight $\alpha$, which usually decreases close to 0 towards the end of training.
%
This means that the entropy bonus $\alpha\mathcal{H}(\pi)$ eventually becomes very small and the task reward is recovered.
%
However, there are certain scenarios where we want to use a fixed $\alpha$ \citep{Haarnoja2018a,eysenbach2021maximum} that always induces a high entropy throughout the training.
%
One natural question is, if $\alpha\mathcal{H}(\pi)$ does not vanish as the training goes, will it possibly hurt the final performance?

Following \citet{Haarnoja2018a}, we use a fixed $\alpha=0.2$ to train SAC, SACZero, and SACLite on the four locomotion tasks in Section~\ref{sec:locomotion}.
%
When drawing training curves, we evaluate a stochastic policy in a deterministic way by taking the mode at each training iteration.
\fi

\section{Entropy Cost Alone Promotes Robustness}
\citet{eysenbach2021maximum} shows (both theoretically and empirically) that MaxEnt RL trains policies that have both dynamics robustness and reward robustness.
In this section, we provide preliminary empirical evidence of the entropy cost alone also resulting in these two types of robustness.

\begin{figure}[t!]
\begin{center}
\resizebox{0.95\columnwidth}{!}{
\begin{tabular}{@{}c@{}c@{}}
    \begin{tabular}{@{}c@{}c}
        \includegraphics[width=0.4\columnwidth]{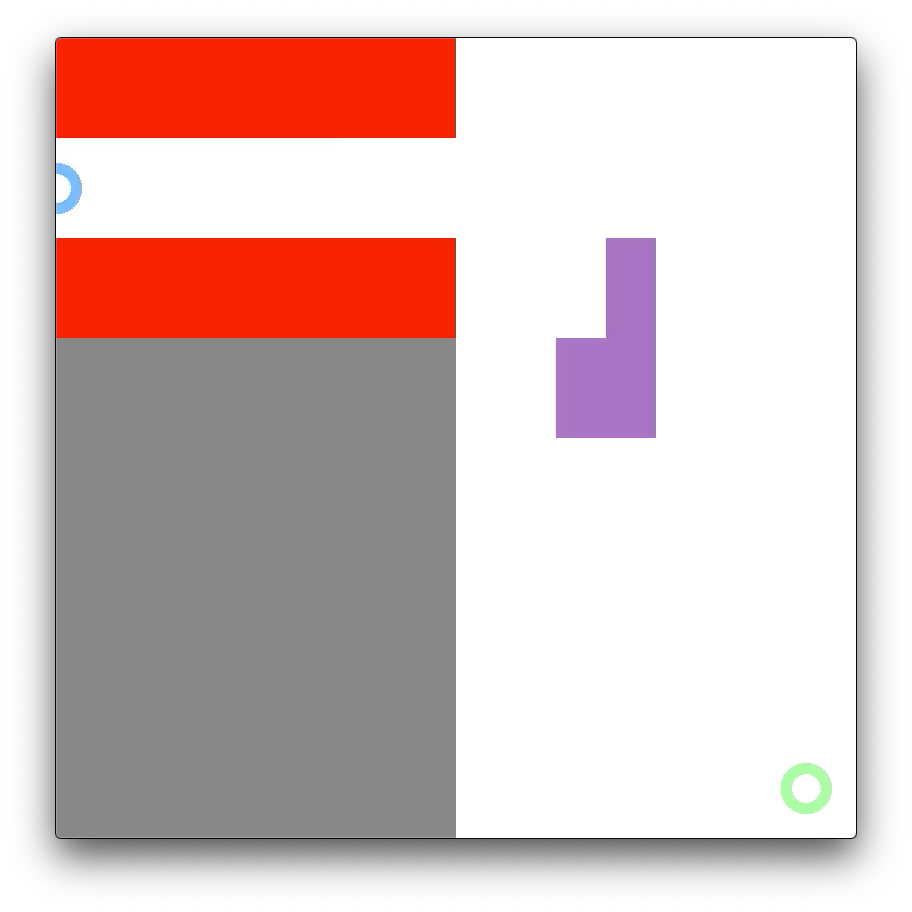}&
        \includegraphics[width=0.4\columnwidth]{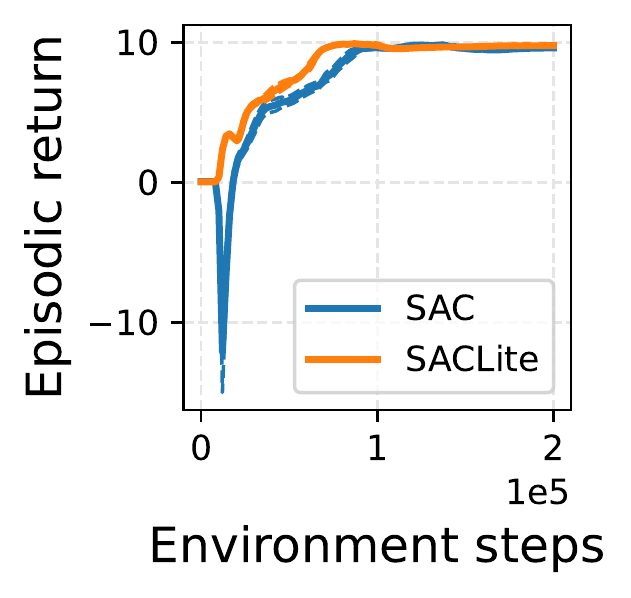}\\
    \end{tabular}&
    \begin{tabular}{@{}r|c@{}}
             & Eval\\
             \hline
             SAC & $9.46$\\
             SACLite & $9.68$\\
    \end{tabular}
\end{tabular}
}
\vspace{-2ex}
\caption{The 2D navigation task (left), its training curves (middle), and the evaluation return (right).
Note that the L-shaped obstacle only appears in evaluation but not in training.
}
\label{fig:toy_navigation}
\end{center}
\vspace{-5ex}
\end{figure}

\textbf{Dynamics robustness}\ \ We first define a similar 2D navigation task (Figure~\ref{fig:toy_navigation}, left) following the one in \citet{eysenbach2021maximum}.
The agent (blue circle) is situated in an $8\times 8$ 2D environment. 
In each episode of up to $50$ steps, the agent starts from the top-left corner to reach the goal (green circle) at the bottom-right corner. 
The reward is defined as the deceased L2 distance to the goal compared to the previous step.
However, the agent will get a penalty of $-1$ every time it enters either of the two red regions.
Thus the optimal return will be roughly $\sqrt{(8-1.5)^2+8^2}=10.3$.
The L-shaped obstacle in purple does \emph{not} exist during training, but will be enabled when evaluating a trained policy, thus adding dynamics perturbations.
The agent has a 2D observation as its current $(x,y)$ coordinate and a 2D action $(dx,dy)\in [-1,1]^2$.
We train SAC and SACLite on this navigation task.
Figure~\ref{fig:toy_navigation} (right) shows the training curves of SAC and SACLite, with SACLite being slightly more sample efficient than SAC.
After this, we enable the L-shaped obstacle on the map and evaluate either trained policy by sampling actions from it. 
Averaged over $500$ episodes, we obtained an episodic return of $9.46$ and $9.68$ for SAC and SACLite, respectively.
This result questions the necessity of the entropy reward for dynamics robustness in this task.

\textbf{Reward robustness}\ \ Next we test if SACLite (trained with the task reward but not the entropy reward) can simultaneously obtain good ``worst-case'' adversarial rewards $\tilde{r}(s,a)=r(s,a)-\alpha\log\pi(a|s)$ as defined in \citet{eysenbach2021maximum}.
We use a fixed $\alpha=0.1$ to ensure that the second term does not fade as training proceeds.
For reference, we also train SAC with the same $\alpha$.
The return gap between the task reward and the adversarial reward is plotted in Figure~\ref{fig:reward_robust_curves}.
It is clear that even trained with the task reward only, SACLite still achieves desirable performance when evaluated in an adversarial setting, showing a large degree of reward robustness.

\section{Discussions, Limitations, and Conclusions}
Through illustrative examples and extensive experiments, we have shown that entropy for regularizing policy evaluation is not ``icing on the cake''.
There is a certain cost involved by adding an entropy reward to the task reward.
We have identified a side effect of the entropy reward which we call reward inflation, inflating/deflating reward function before but not after episode terminations.
Reward inflation could obscure or completely change the original MDP in an episodic setting.
%
%Even for an infinite horizon, reward inflation might still exist in practice due to the initial bias in functionally approximated Q values.
Even for an infinite horizon,
the non-stationarity of Q values caused by a decaying entropy reward also poses a challenge to policy optimization.
Moreover, the fact that SACZero is worse than SACLite on some tasks demonstrates that even after removing reward inflation, 
the entropy reward could still obscure the task reward by changing the relative importance of every transition.
With a tunable $\alpha$, the entropy reward usually vanishes after some time and MaxEnt RL reduces to standard RL.
However, by then $\alpha$ is very small and exploration can hardly be done. 
The policy may have got stuck in a local optimum.

We have also shown that the properties of good exploration, training convergence and stability, and policy robustness of MaxEnt RL, result more from entropy regularizing policy improvement than from entropy regularizing policy evaluation.
A theoretical explanation of this might be helpful for better understanding our results in the future.

It is impractical for us to evaluate all continuous control RL tasks available out there.
Our evaluation has selected representative tasks that can be approached by SAC reasonably while still posing challenges.
Our goal is to at least bring some side effects of the entropy reward to RL practitioners' attention.
We believe that this work can shed some light on how to better harness entropy as an intrinsic reward to improve entropy-regularized RL training in practice.

\ifdefined\isaccepted
\section*{Acknowledgements}
We thank the Horizon AIDI platform team for computational infrastructure support, and the group members for help discussions.
\else\fi

\bibliography{ICML2022_closer_look_entropy}

\begin{thebibliography}{25}
\providecommand{\natexlab}[1]{#1}
\providecommand{\url}[1]{\texttt{#1}}
\expandafter\ifx\csname urlstyle\endcsname\relax
  \providecommand{\doi}[1]{doi: #1}\else
  \providecommand{\doi}{doi: \begingroup \urlstyle{rm}\Url}\fi

\bibitem[Abdolmaleki et~al.(2018)Abdolmaleki, Springenberg, Tassa, Munos,
  Heess, and Riedmiller]{abdolmaleki2018maximum}
Abdolmaleki, A., Springenberg, J.~T., Tassa, Y., Munos, R., Heess, N., and
  Riedmiller, M.
\newblock Maximum a posteriori policy optimisation.
\newblock In \emph{ICLR}, 2018.

\bibitem[Ahmed et~al.(2019)Ahmed, Roux, Norouzi, and
  Schuurmans]{ahmed2019understanding}
Ahmed, Z., Roux, N.~L., Norouzi, M., and Schuurmans, D.
\newblock Understanding the impact of entropy on policy optimization.
\newblock In \emph{ICML}, 2019.

\bibitem[Badia et~al.(2020)Badia, Sprechmann, Vitvitskyi, Guo, Piot,
  Kapturowski, Tieleman, Arjovsky, Pritzel, Bolt, and Blundell]{Badia2020}
Badia, A.~P., Sprechmann, P., Vitvitskyi, A., Guo, Z.~D., Piot, B.,
  Kapturowski, S., Tieleman, O., Arjovsky, M., Pritzel, A., Bolt, A., and
  Blundell, C.
\newblock Never give up: Learning directed exploration strategies.
\newblock In \emph{ICLR}, 2020.

\bibitem[Bohez et~al.(2019)Bohez, Abdolmaleki, Neunert, Buchli, Heess, and
  Hadsell]{Bohez2019}
Bohez, S., Abdolmaleki, A., Neunert, M., Buchli, J., Heess, N., and Hadsell, R.
\newblock Value constrained model-free continuous control.
\newblock \emph{arXiv}, 2019.

\bibitem[Brockman et~al.(2016)Brockman, Cheung, Pettersson, Schneider,
  Schulman, Tang, and Zaremba]{Brockman2016}
Brockman, G., Cheung, V., Pettersson, L., Schneider, J., Schulman, J., Tang,
  J., and Zaremba, W.
\newblock Openai gym, 2016.

\bibitem[Burda et~al.(2018)Burda, Edwards, Storkey, and
  Klimov]{burda2018exploration}
Burda, Y., Edwards, H., Storkey, A., and Klimov, O.
\newblock Exploration by random network distillation.
\newblock In \emph{ICLR}, 2018.

\bibitem[Eysenbach \& Levine(2021)Eysenbach and Levine]{eysenbach2021maximum}
Eysenbach, B. and Levine, S.
\newblock Maximum entropy rl (provably) solves some robust rl problems.
\newblock \emph{arXiv}, 2021.

\bibitem[Eysenbach et~al.(2019)Eysenbach, Gupta, Ibarz, and
  Levine]{Eysenbach2019}
Eysenbach, B., Gupta, A., Ibarz, J., and Levine, S.
\newblock Diversity is all you need: Learning skills without a reward function.
\newblock In \emph{ICLR}, 2019.

\bibitem[Fujimoto et~al.(2018)Fujimoto, van Hoof, and
  Meger]{fujimoto2018addressing}
Fujimoto, S., van Hoof, H., and Meger, D.
\newblock Addressing function approximation error in actor-critic methods.
\newblock In \emph{ICML}, 2018.

\bibitem[Gehring et~al.(2021)Gehring, Synnaeve, Krause, and
  Usunier]{Gehring2021}
Gehring, J., Synnaeve, G., Krause, A., and Usunier, N.
\newblock Hierarchical skills for efficient exploration.
\newblock In \emph{NeurIPS}, 2021.

\bibitem[Haarnoja et~al.(2017)Haarnoja, Tang, Abbeel, and
  Levine]{haarnoja2017softq}
Haarnoja, T., Tang, H., Abbeel, P., and Levine, S.
\newblock Reinforcement learning with deep energy-based policies.
\newblock In \emph{ICML}, 2017.

\bibitem[Haarnoja et~al.(2018)Haarnoja, Zhou, Hartikainen, Tucker, Ha, Tan,
  Kumar, Zhu, Gupta, Abbeel, and Levine]{Haarnoja2018}
Haarnoja, T., Zhou, A., Hartikainen, K., Tucker, G., Ha, S., Tan, J., Kumar,
  V., Zhu, H., Gupta, A., Abbeel, P., and Levine, S.
\newblock Soft actor-critic algorithms and applications.
\newblock \emph{arXiv}, abs/1812.05905, 2018.

\bibitem[Kingma \& Ba(2015)Kingma and Ba]{kingma2017adam}
Kingma, D.~P. and Ba, J.
\newblock Adam: A method for stochastic optimization.
\newblock In \emph{ICLR}, 2015.

\bibitem[Lillicrap et~al.(2016)Lillicrap, Hunt, Pritzel, Heess, Erez, Tassa,
  Silver, and Wierstra]{Lillicrap2016}
Lillicrap, T.~P., Hunt, J.~J., Pritzel, A., Heess, N., Erez, T., Tassa, Y.,
  Silver, D., and Wierstra, D.
\newblock Continuous control with deep reinforcement learning.
\newblock In \emph{ICLR}, 2016.

\bibitem[Mnih et~al.(2016)Mnih, Badia, Mirza, Graves, Lillicrap, Harley,
  Silver, and Kavukcuoglu]{mnih2016asynchronous}
Mnih, V., Badia, A.~P., Mirza, M., Graves, A., Lillicrap, T.~P., Harley, T.,
  Silver, D., and Kavukcuoglu, K.
\newblock Asynchronous methods for deep reinforcement learning.
\newblock In \emph{ICML}, 2016.

\bibitem[Nachum et~al.(2017)Nachum, Norouzi, Xu, and
  Schuurmans]{nachum2017bridging}
Nachum, O., Norouzi, M., Xu, K., and Schuurmans, D.
\newblock Bridging the gap between value and policy based reinforcement
  learning.
\newblock In \emph{NeurIPS}, 2017.

\bibitem[Pathak et~al.(2017)Pathak, Agrawal, Efros, and
  Darrell]{pathak2017curiositydriven}
Pathak, D., Agrawal, P., Efros, A.~A., and Darrell, T.
\newblock Curiosity-driven exploration by self-supervised prediction.
\newblock In \emph{ICML}, 2017.

\bibitem[Plappert et~al.(2018)Plappert, Andrychowicz, Ray, McGrew, Baker,
  Powell, Schneider, Tobin, Chociej, Welinder, Kumar, and
  Zaremba]{plappert2018multigoal}
Plappert, M., Andrychowicz, M., Ray, A., McGrew, B., Baker, B., Powell, G.,
  Schneider, J., Tobin, J., Chociej, M., Welinder, P., Kumar, V., and Zaremba,
  W.
\newblock Multi-goal reinforcement learning: Challenging robotics environments
  and request for research, 2018.

\bibitem[Ray et~al.(2019)Ray, Achiam, and Amodei]{Ray2019}
Ray, A., Achiam, J., and Amodei, D.
\newblock {Benchmarking Safe Exploration in Deep Reinforcement Learning}.
\newblock 2019.

\bibitem[Schulman et~al.(2017)Schulman, Wolski, Dhariwal, Radford, and
  Klimov]{schulman2017proximal}
Schulman, J., Wolski, F., Dhariwal, P., Radford, A., and Klimov, O.
\newblock Proximal policy optimization algorithms.
\newblock \emph{arXiv}, 2017.

\bibitem[Schulman et~al.(2018)Schulman, Chen, and
  Abbeel]{schulman2018equivalence}
Schulman, J., Chen, X., and Abbeel, P.
\newblock Equivalence between policy gradients and soft q-learning.
\newblock \emph{arXiv}, 2018.

\bibitem[Tassa et~al.(2020)Tassa, Tunyasuvunakool, Muldal, Doron, Liu, Bohez,
  Merel, Erez, Lillicrap, and Heess]{tassa2020dmcontrol}
Tassa, Y., Tunyasuvunakool, S., Muldal, A., Doron, Y., Liu, S., Bohez, S.,
  Merel, J., Erez, T., Lillicrap, T., and Heess, N.
\newblock dm\_control: Software and tasks for continuous control, 2020.

\bibitem[Vieillard et~al.(2020)Vieillard, Kozuno, Scherrer, Pietquin, Munos,
  and Geist]{vieillard2020leverage}
Vieillard, N., Kozuno, T., Scherrer, B., Pietquin, O., Munos, R., and Geist, M.
\newblock Leverage the average: an analysis of kl regularization in rl.
\newblock In \emph{NeurIPS}, 2020.

\bibitem[Yu et~al.(2022)Yu, Xu, and Zhang]{Yu2022seditor}
Yu, H., Xu, W., and Zhang, H.
\newblock Towards safe reinforcement learning with a safety editor policy.
\newblock \emph{arXiv}, 2022.

\bibitem[Ziebart(2010)]{Ziebart2010}
Ziebart, B.~D.
\newblock \emph{Modeling purposeful adaptive behavior with the principle of
  maximum causal entropy}.
\newblock PhD thesis, Carnegie Mellon University, 2010.

\end{thebibliography}
\bibliographystyle{icml2021}

\clearpage
\appendix

\section{Hyperparameters}
\label{app:hyperparameters}
Here, we briefly list the most important hyperparameters for each experiment in 
the paper.
These hyperparameter values were selected as the best practices on these tasks in our daily research, 
and they generally ensure that SAC is able to achieve reasonably good performance.
For more details, please refer to our released code (Section~\ref{sec:introduction}).

\subsection{Common}
We first describe the common hyperparameters that are adopted in each experiment. 
They are fixed across experiments unless otherwise stated.
We initially collect 10k environment steps in the replay buffer before the training starts.
The smoothing coefficient of the target Q network is $5\times 10^{-3}$, and the soft target update happens every training iteration.
The training interval is 1 environment step.
The hidden nonlinear activations of all networks are ReLU.
The discount factor $\gamma$ is set to $0.99$.
We use the Adam optimizer~\citep{kingma2017adam} with $\beta_1=0.9$, $\beta_2=0.999$, and $\epsilon=10^{-7}$ for training.

\subsection{\textsc{SimpleChain}}
Only one actor is used for collecting experiences. 
Both the actor and critic networks have only one hidden layer of size $100$.
The initial number of environment steps collected in the replay buffer is 5k.
The total number of training environment steps is 50k which is also the replay buffer size.
The entropy target for each action dimension is $-1$.
We use a learning rate of $10^{-4}$ with a mini-batch size of $256$.

\subsection{Manipulation}
\begin{figure}[!htb]
\begin{center}
\resizebox{\columnwidth}{!}{
    \begin{tabular}{@{}c@{}c@{}c@{}c@{}}
        \includegraphics[width=0.4\columnwidth]{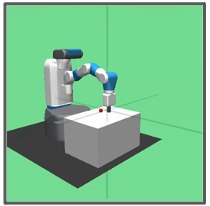}&
        \includegraphics[width=0.43\columnwidth]{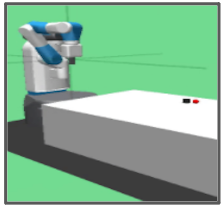}&
        \includegraphics[width=0.4\columnwidth]{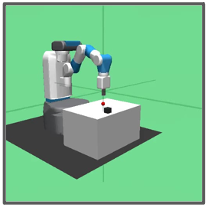}&
        \includegraphics[width=0.4\columnwidth]{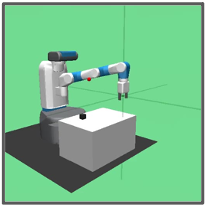}\\
        \textsc{FetchReach} & \textsc{FetchSlide} & \textsc{FetchPush} & \textsc{FetchPickAndPlace}\\
    \end{tabular}
}
\end{center}
\end{figure}
The hyperparameters of this experiment closely followed those in \citet{plappert2018multigoal}.
We use 38 actors to collect experiences in parallel.
Both the actor and critic networks have three hidden layers of sizes $(256,256,256)$.
The discount factor is set to $0.98$.
The total number of training environment steps is 10M, and the replay buffer size is $38\times 20\text{k}=0.96$M.
The smoothing coefficient of the target Q network is $0.05$, and the soft target update happens every $40$ training
iterations.
The training interval is $47.5$ ($1900/40$) environment steps. 
We set the entropy target for each action dimension to about $-0.916$, which roughly assumes that the target action distribution has a probability 
mass concentrated on $\frac{1}{5}$ of the support $[-1,1]$.
We use a learning rate of $10^{-3}$ with a mini-batch size of $4864$.
We normalize the task reward by its running average and standard deviation, and clip the normalized reward to $[-1,1]$.

\subsection{Box2D}
\begin{figure}[!htb]
\begin{center}
\resizebox{\columnwidth}{!}{
    \begin{tabular}{@{}cccc@{}}
        \includegraphics[width=0.45\columnwidth]{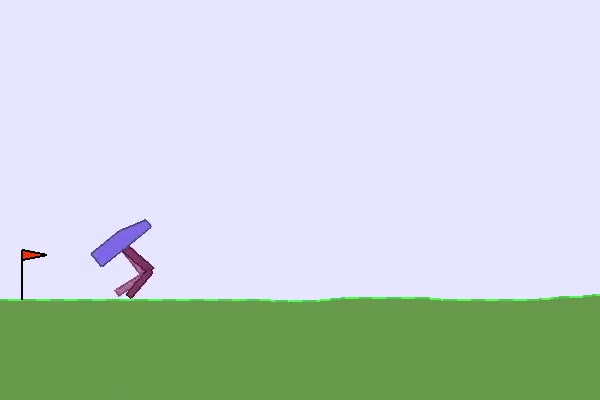}&
        \includegraphics[width=0.45\columnwidth]{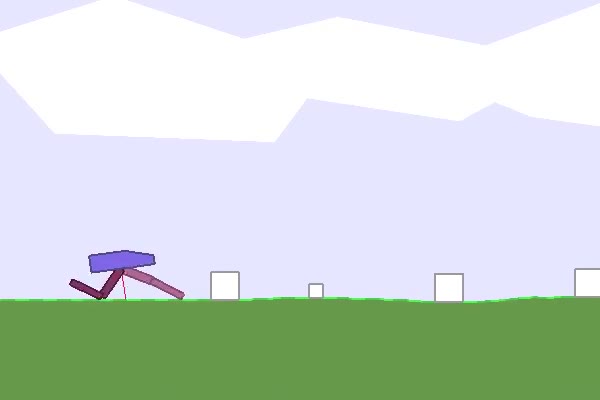}&
        \includegraphics[width=0.45\columnwidth]{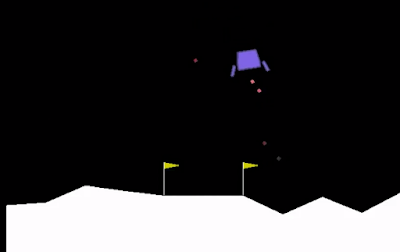}&
        \includegraphics[width=0.45\columnwidth]{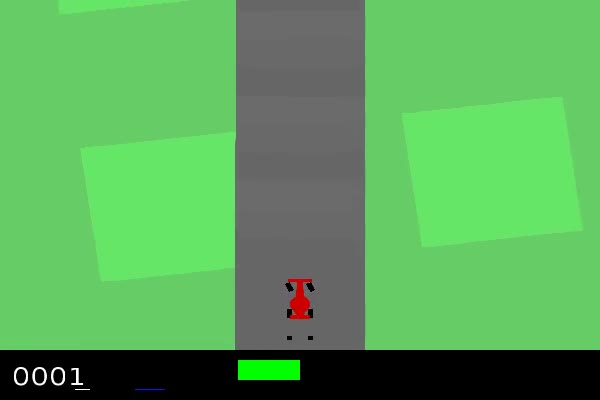}\\
        \textsc{BipedalWalker} & \textsc{BipedalWalkerHardcore} & \textsc{LunarLanderContinuous} & \textsc{CarRacing}\\
    \end{tabular}
}
\end{center}
\end{figure}
\textsc{BipedalWalker} and \textsc{BipedalWalkerHardcore}\ \ We use 32 actors to collect experiences in parallel.
Both the actor and critic networks have two hidden layers of sizes $(256,256)$. 
The total number of training environment steps is 5M, and the replay buffer size is $32\times 100\text{k}=3.2\text{M}$.
The training interval is $128$ environment steps.
We set the entropy target for each action dimension to about $-1.609$, which roughly assumes that the target action distribution has a probability mass concentrated on $\frac{1}{10}$ of the support $[-1,1]$.
We use a learning rate of $5\times 10^{-4}$ and a mini-batch size of $4096$.

\textsc{LunarLanderContinuous}\ \ We use only one actor for collecting experiences. 
Both the actor and critic networks have two hidden layers of sizes $(256,256)$.
The total number of training environment steps is 100k, and the replay buffer has the same size.
The entropy target is set in the same way as \textsc{BipedalWalker}.
We use a learning rate of $10^{-4}$ with a mini-batch size of $256$.

\textsc{CarRacing}\ \ We use 16 actors to collect experiences in parallel.
Four gray-scale contiguous frames are stacked as the input observation for each step.
The initial number of environment steps collected in the replay buffer is 50k.
Both the actor and critic networks have a CNN configured as $[(32, 8, 4), (64, 4, 2), (64, 3, 1)]$, where each triplet represents channels, kernel size, and stride. 
The CNN is followed by two hidden layers of sizes $(256,256)$.
The total number of training environment steps is 10M, and the replay buffer size is $16\times 100\text{k}=1.6\text{M}$.
The training interval is $80$ environment steps.
The entropy target is set in the same way as \textsc{BipedalWalker}.
We use a learning rate of $3\times 10^{-4}$ with a mini-batch size of $256$.
Additionally, 7-step TD learning is employed to speed up training.

We normalize the task reward of every Box2D task by its running average and standard deviation, and clip the normalized reward to $[-5,5]$.

\subsection{MuJoCo locomotion}
\begin{figure}[!htb]
\begin{center}
\resizebox{\columnwidth}{!}{
\begin{tabular}{@{}c@{}c@{}c@{}c@{}}
    \includegraphics[width=0.34\columnwidth]{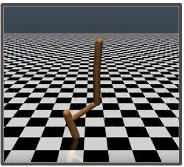}&
    \includegraphics[width=0.3\columnwidth]{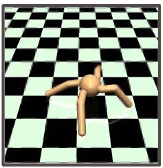}&
    \includegraphics[width=0.3\columnwidth]{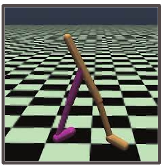}&
    \includegraphics[width=0.31\columnwidth]{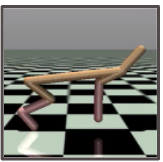}\\
    \textsc{Hopper}& \textsc{Ant}& \textsc{Walker2d}& \textsc{HalfCheetah}\\
\end{tabular}
}
\end{center}
\end{figure}
The hyperparameters of this experiment closely followed those in \citet{Haarnoja2018}.
We use only one actor to collect experiences. 
Both the actor and critic networks have two hidden layers of sizes $(256,256)$.
The total number of training environment steps is 3M, and the replay buffer size is 1M.
The initial number of environment steps collected is the replay buffer is 50k.
The entropy target for each action dimension is $-1$.
We use a learning rate of $3\times 10^{-4}$ with a mini-batch size of $256$.

\subsection{DM control suite}
\begin{figure}[!htb]
\begin{center}
\resizebox{\columnwidth}{!}{
\begin{tabular}{@{}ccc@{}}
    \includegraphics[width=0.33\columnwidth]{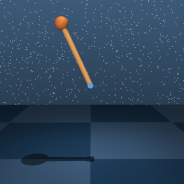}&
    \includegraphics[width=0.33\columnwidth]{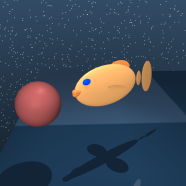}&
    \includegraphics[width=0.33\columnwidth]{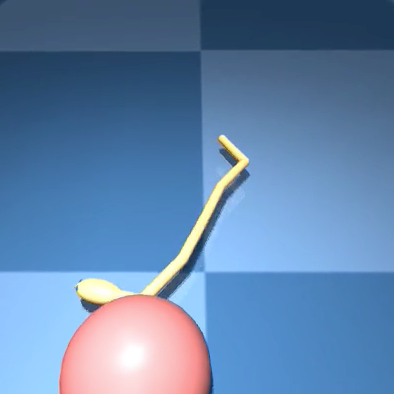}\\
    \textsc{pendulum} & \textsc{fish} & \textsc{swimmer}\\
    \includegraphics[width=0.33\columnwidth]{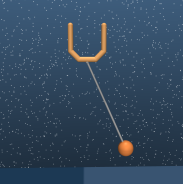}&
    \includegraphics[width=0.33\columnwidth]{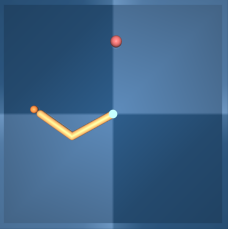}&
    \includegraphics[width=0.33\columnwidth]{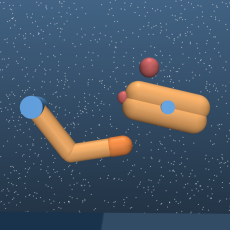}\\
    \textsc{ball\_in\_cup} & \textsc{reacher} & \textsc{finger}\\    
\end{tabular}
}
\end{center}
\end{figure}
We use 10 actors to collect experiences in parallel. 
Each action is repeated twice at every step.
Both the actor and critic networks have two hidden layers of sizes $(256,256)$.
The total number of training environment steps is 3M, and the replay buffer size is $10\times 100\text{k}=1\text{M}$.
The training interval is 10 environment steps.
The entropy target is set in the same way as the manipulation tasks.
We use a learning rate of $3\times10^{-4}$ with a mini-batch size of $256$.
Additionally, 5-step TD learning is employed to speed up training.
We didn't normalize the task reward as it is already in a good range of $[0,1]$.

\subsection{Multi-objective RL}
\begin{figure}[!htb]
\begin{center}
\resizebox{\columnwidth}{!}{
\begin{tabular}{@{}c@{}c@{}c@{}}
    \includegraphics[width=0.4\columnwidth]{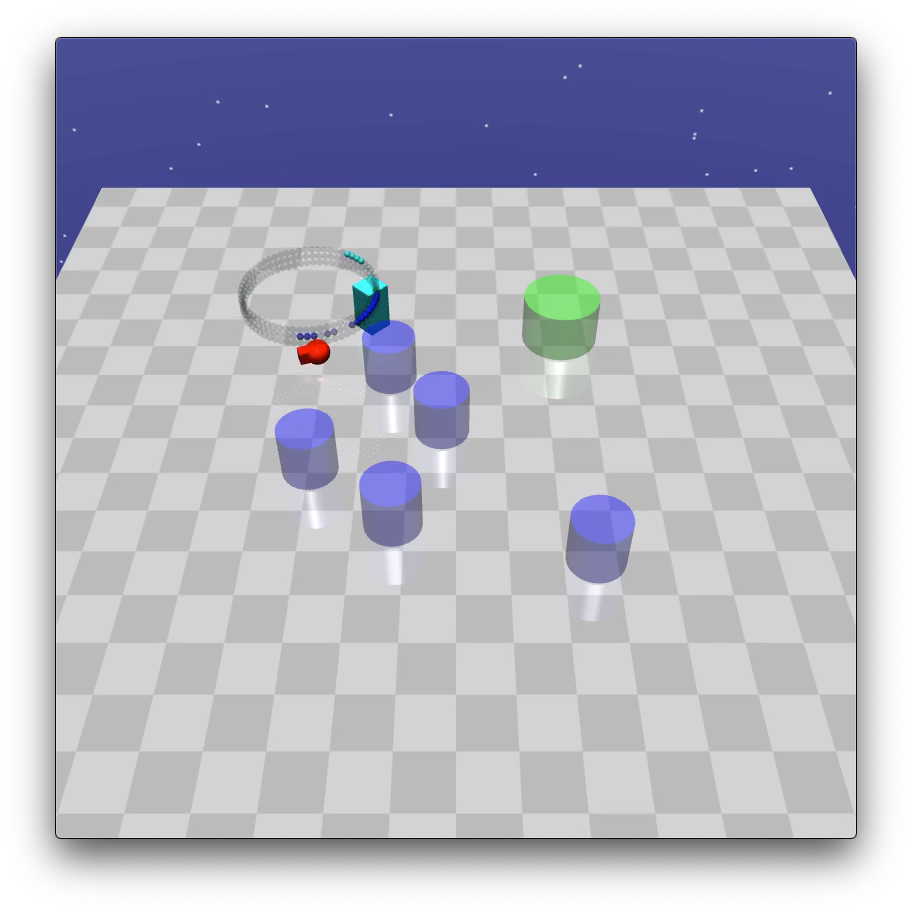}&
    \includegraphics[width=0.4\columnwidth]{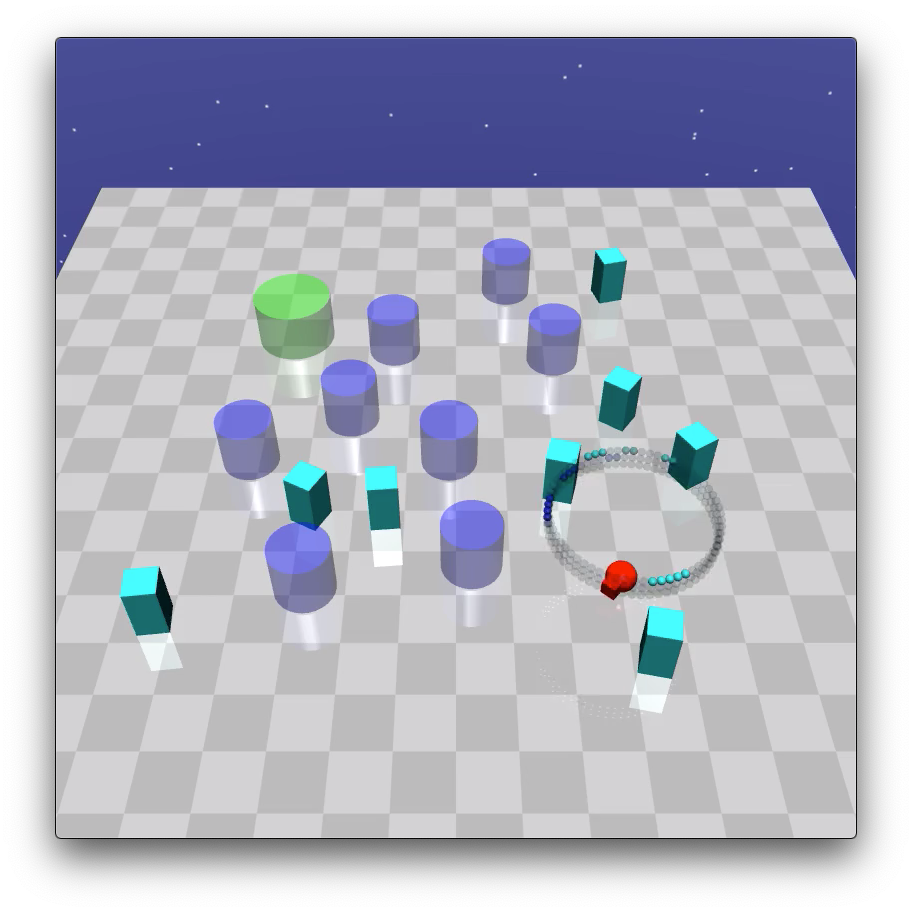}&
    \includegraphics[width=0.4\columnwidth]{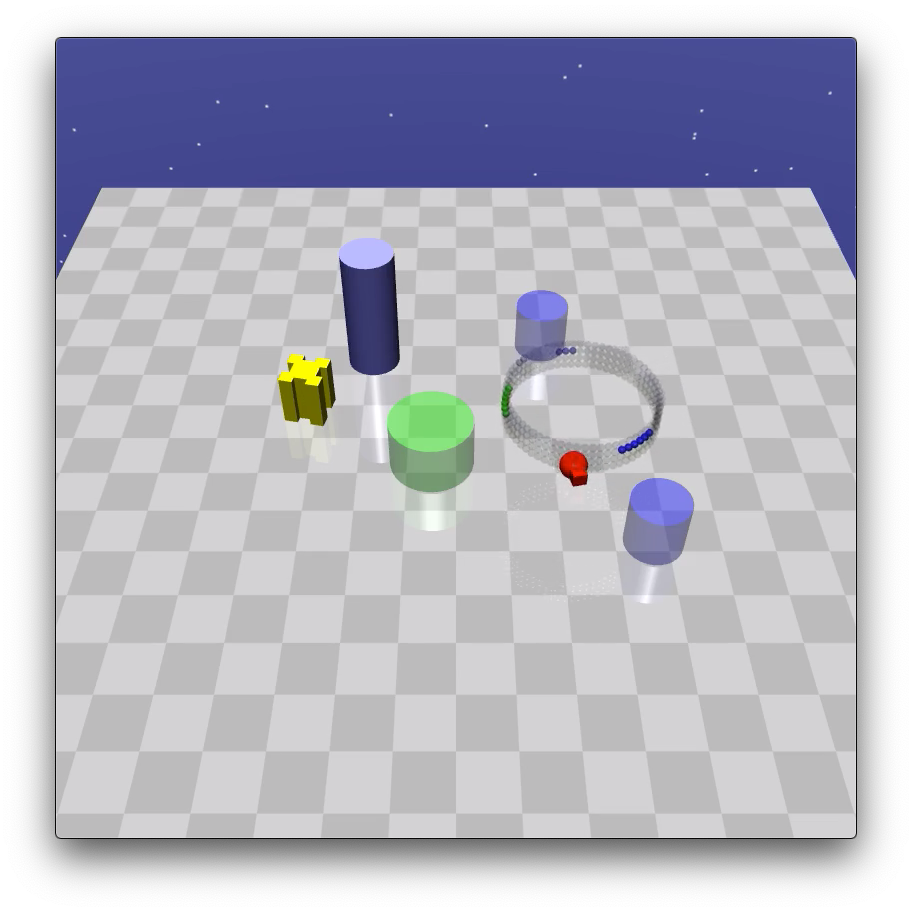}\\
    \textsc{PointGoal1} & \textsc{PointGoal2} & \textsc{PointPush1}\\
    \includegraphics[width=0.4\columnwidth]{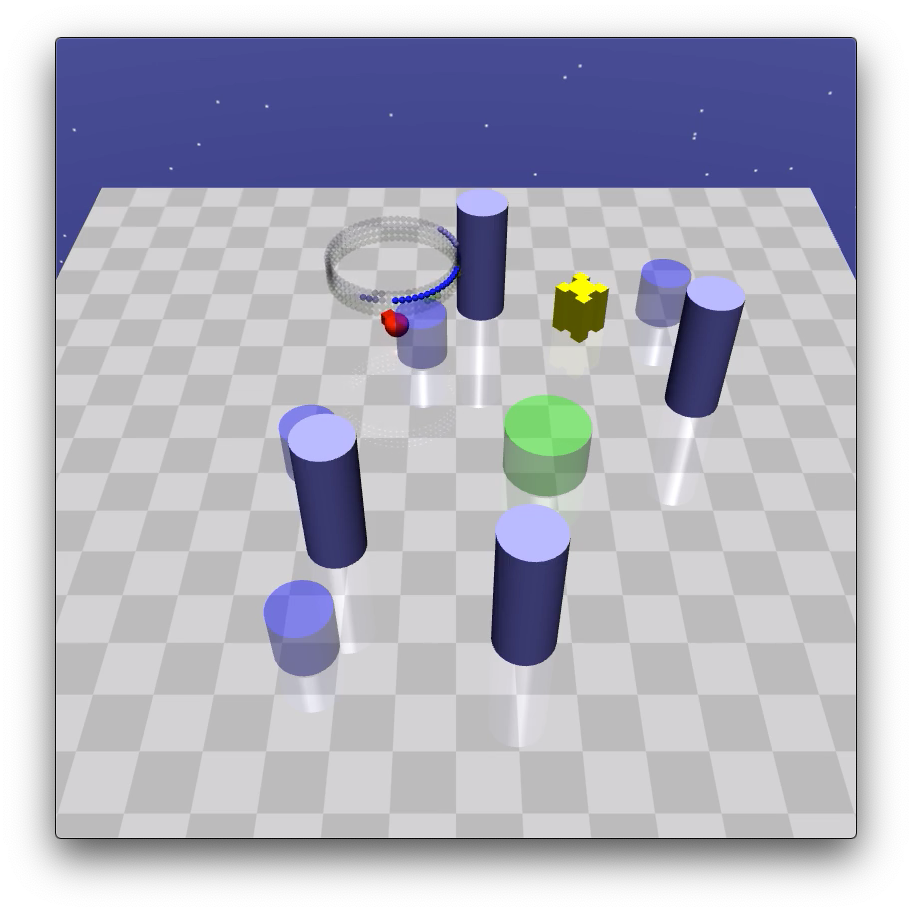}&
    \includegraphics[width=0.4\columnwidth]{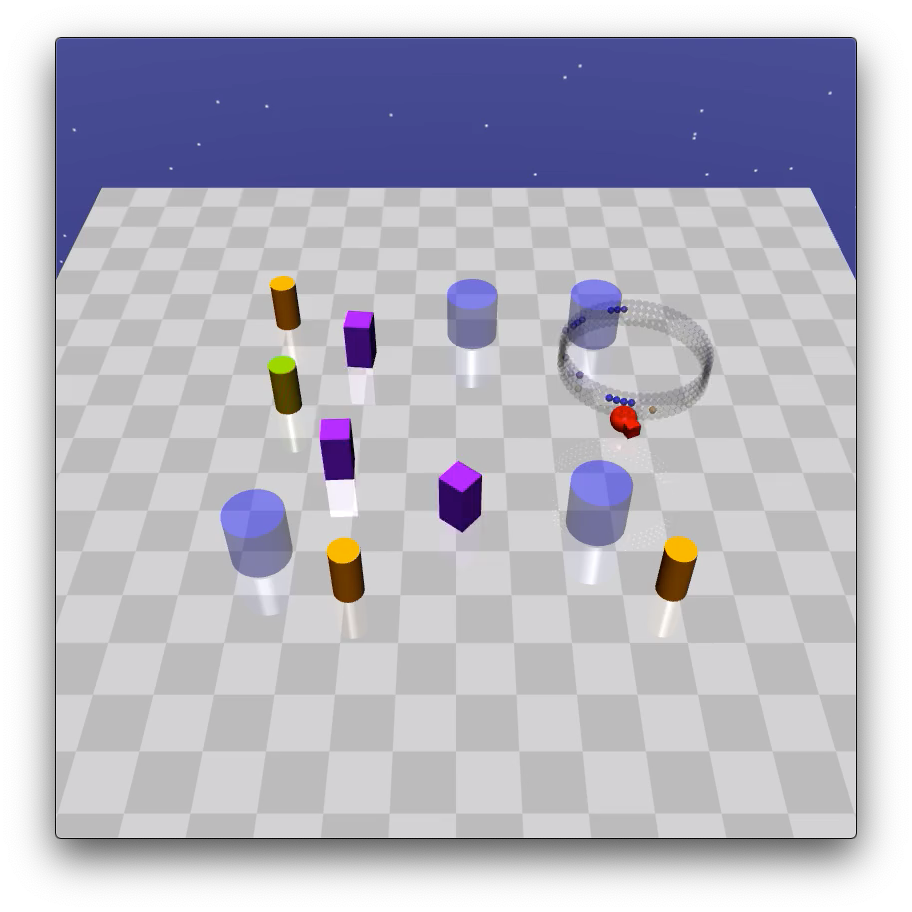}&
    \includegraphics[width=0.4\columnwidth]{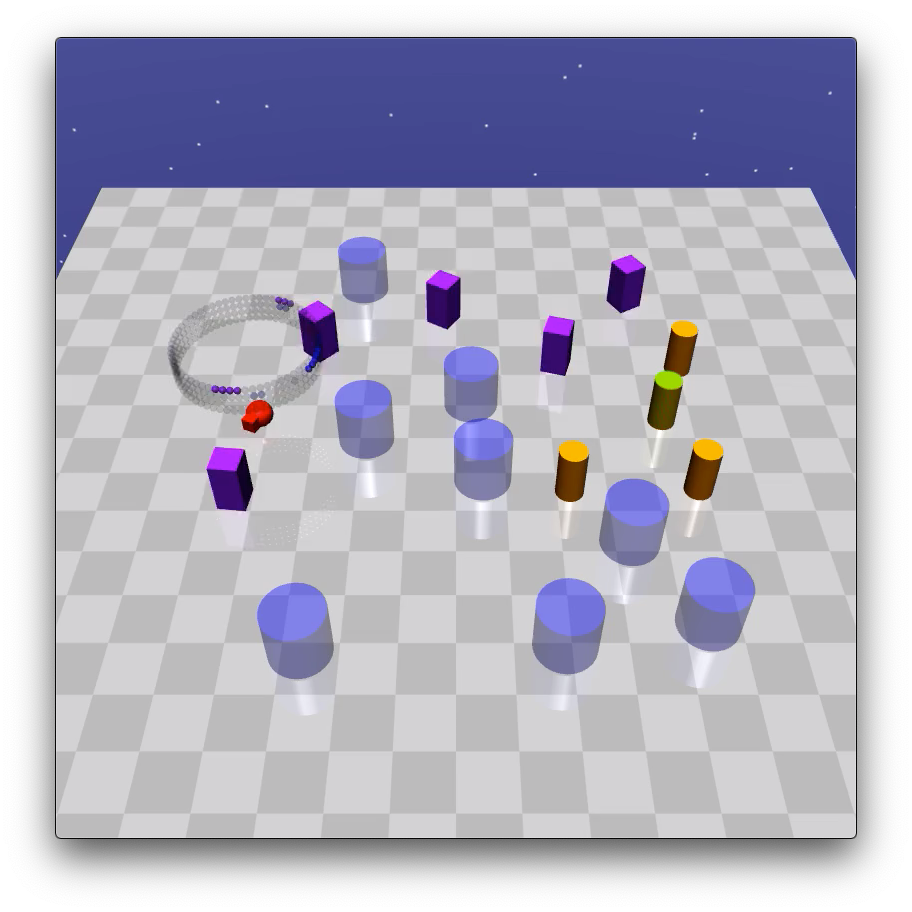}\\
    \textsc{PointPush2} & \textsc{PointButton1} & \textsc{PointButton2}\\    
\end{tabular}
}
\end{center}
\end{figure}
Following
\ifdefined\isaccepted
\citet{Yu2022seditor},
\else
\citet{Yu2022seditorAnonymous},
\fi
we customized the Safety Gym~\citep{Ray2019} environment to enable a natural lidar of 64 bins to replace the default 
pseudo lidar of 16 bins for generating observations, as the natural lidar gives more information about the object shapes in the environment.
We found that rich shape information is necessary for the agent to achieve a low constraint threshold.
A separate lidar vector of length 64 is produced for each obstacle type or goal.
All lidar vectors and a vector of the robot status are concatenated together to produce a flattened observation vector.

We use 32 actors to collect experiences in parallel.
Since the two Button tasks contain moving gremlins to be avoided, we stack four observations as the input observation for each step.
This frame stacking is adopted for all six tasks.
Both the actor and critic networks have three hidden layers of sizes $(256,256,256)$.
The total number of training environment steps is 5M, and the replay buffer size is $32\times 50\text{k}=1.6\text{M}$.
The training interval is $160$ environment steps.
The entropy target is set in the same way as \textsc{BipedalWalker}.
We use a learning rate of $0.01$ for the Lagrangian method and $3\times 10^{-4}$ for the rest of the algorithm, with a mini-batch size of $1024$.
Additionally, 7-step TD learning is employed to speed up training.
Following~\citet{Bohez2019}, when using the Lagrangian multiplier $\lambda>0$ to combine the two Q values, we use the normalized weights $\frac{1}{1+\lambda}$ and $\frac{\lambda}{1+\lambda}$.
This makes the combined Q value bounded.
We normalize the task reward vector by its running average and standard deviation, and clip the normalized vector to $[-10,10]$.

%%%%%%%%%%%%%%%%%%%%%%%%%%%%%%%%%%%%%%%%%%%%%%%%%%%%%%%%%%%%%%%%%%%%%%%%%%%%%%%
%%%%%%%%%%%%%%%%%%%%%%%%%%%%%%%%%%%%%%%%%%%%%%%%%%%%%%%%%%%%%%%%%%%%%%%%%%%%%%%

\end{document}